Measuring Lexical Diversity in Texts: The Twofold Length Problem

Yves Bestgen

Université catholique de Louvain

**Abstract**

The impact of text length on the estimation of lexical diversity has captured the attention of the scientific community for more than a century. Numerous indices have been proposed, and many studies have been conducted to evaluate them, but the problem remains. This methodological review provides a critical analysis not only of the most commonly used indices in language learning studies, but also of the length problem itself, as well as of the methodology for evaluating the proposed solutions. The analysis of three datasets of English language-learners' texts revealed that indices that reduce all texts to the same length using a probabilistic or an algorithmic approach solve the length dependency problem; however, all these indices failed to address the second problem, which is their sensitivity to the parameter that determines the length to which the texts are reduced. The paper concludes with recommendations for optimizing lexical diversity analysis.

**Measuring Lexical Diversity in Texts: The Twofold Length Problem**

Lexical diversity (LD) is defined as "the range and variety of vocabulary deployed in a text" (McCarthy & Jarvis, 2007, p. 459). It is the subject of a significant amount of research in literature, political text analysis and, particularly, in studies of language learning and impairment (Jarvis, 2013; Malvern et al., 2004). At first glance, its evaluation appears to be quite easy, as counting the number of different words (or types) in a text appears to be sufficient. In reality, the task is far from being simple because the length of the text has an impact on the number of types it contains. The most obvious solution, which consists of dividing the number of types by the total number of words (or tokens), is not the solution due to the nonlinear relationship between the numbers of types and tokens (Tweedie & Baayen, 1998). This long-standing observation has led to a proliferation of studies aimed at proposing indices that are supposedly insensitive to text length, evaluating them in various situations, and attempting to determine which text size can be analyzed using which index (e.g., Fergadiotis et al., 2013, 2015; Hess et al., 1989; Jarvis, 2002; Jarvis & Hashimoto, 2021; Koizumi & In'nami, 2012; Lu, 2012; Malvern et al., 2004; McCarthy & Jarvis, 2007, 2010; Nasseri & Thompson, 2021; Treffers-Daller, 2013; Zenker & Kyle, 2021). This work has led to contradictory conclusions. In particular, some indices that were once considered insensitive to length have been shown not to be so in new studies, while other indices have followed the opposite path. Similarly, the solution of reserving certain indices for texts of certain lengths have not proven to be more reliable.

The first justification for this contribution is that many of these works are plagued by methodological problems that make their conclusions incorrect, and that these problems can be remedied. Another major limitation of these works is that they have neglected a second problem encountered by the indices that attempt to control the text length dependency by measuring LD in text segments in which length is determined by a parameter. The impact of this parameter is rarely evaluated. Finally, the analysis of the logical-mathematical properties of the indices has been neglected, thus preventing an accurate view of their most important features and the relationships among them. For these three reasons, this methodological review proposes a classification of indices based on their logical-mathematical relationship, a critical discussion of the twofold length problem and of the methods for evaluating it, better statistical indicators of the existence of a problem, and an empirical analysis of three datasets. These datasets consist of written argumentative essays and monologues produced in a testing context, a context in which LD has often been used as a predictor of L2 proficiency. It concludes with recommendations for optimizing LD analyses, particularly in the domain of language learning.

Before addressing these points, it is essential to emphasize that the problem of the sensitivity of LD indices to text length is only one of the difficulties in measuring LD. First, the number of types that are present in a text is only the most obvious dimension of LD. Other features of a text's vocabulary, such as the greater or lesser semantic similarity of the words present, are also relevant (Kyle et al., 2021). These features were studied in great detail a decade ago (Jarvis, 2013), and work is currently being done in an attempt to resolve this issue (Jarvis, 2017; Kyle et al., 2021). In the meantime, the use of indices based on the number of types and tokens has persisted. Second, the definition of a type is not straightforward, as it can be operationalized in different ways, such as spelling forms, lemmas, or word families (Jarvis & Hashimoto, 2021; Treffers-Daller et al., 2018). Regardless of the most appropriate operationalization, length-insensitive LD indices are needed.

### Classification of the Main LD Indices Used in Studies of Language Learning

The first step in any LD study in the domain of language learning is to select one or several indices. The two main criteria employed are insensitivity to text length and sensitivity to differences in language proficiency. As Jarvis (2013) pointed out, this second criterion allows for the selection of effective, yet invalid, indices because the reasons that these indices are effective are not related to LD, but to other factors.

A third criterion is worth considering, namely the logical-mathematical relationships among these indices (Tweedie & Baayen, 1998), which originated in the way in which they attempt to correct for the length effect. Using this criterion enables the use of indices from different categories to cover different points of view of LD or to select indices only from the category that appears to be the most relevant. This is the criterion used in this section. The symbols used are $N$ for the number of tokens in a text, $V$ for the number of types (vocabulary), $s$ for the number of samples extracted from a text and $n$ for the number of tokens in a sample, a segment, or a moving window.

The main indices used to analyze LD in the language-learning domain[1] can be divided into two broad categories: indices that take a global versus those that adopt a local view of LD. Global indices estimate LD by analyzing repetitions in the entire text by treating it as "a bag of words". Local indices also provide an LD score for the text, but on the basis of text segments; that is, contiguous word sequences. They are thus more respectful of the true nature of a text, which is sequential[2] (Tweedie & Baayen, 1998). These indices also have the advantage of being able to be used to track the fluctuation of LD in a text, a characteristic of LD seldom studied in language learning. Compared to global indices, local indices only have partial knowledge of the lexical repetitions in a text because this knowledge is limited to the length of the segments. The consequence is that local indices do not treat the text as a whole (Jarvis, 2013; Malvern et al., 2004). If the conclusion of an essay repeats the introduction extensively, the local approach will miss this, while the global approach will not.

### Indices Taking a Global View of LD

This category includes indices that are based on the full set of types and tokens that a text contains. It can be divided into two subcategories.

#### *Type-token Ratio (TTR) and Nonlinear Transformations of the Number of Types and Tokens*

The archetype in this category is TTR ($V/N$), the sensitivity to text length of which is well established. Accordingly, many transformations of the number of types and tokens have been proposed in an attempt to remove this length effect. The most frequently used in language-learning studies are Guiraud's R, ($V/\sqrt{N}$), Herdan's C, ($\log V/\log N$), and Maas' a, ($\sqrt{(\log N - \log V)/(\log N \times \log N)}$).

#### *Random Sampling and Theoretical Distribution*

The indices in this category estimate the number of types the text would contain if it were reduced to a given length. These indices are based on the urn model (Baayen, 2001). In this model, the urn contains all the tokens in the text, and a reduced image of the text is obtained by randomly extracting a certain number of tokens. The more often this draw is repeated, the more faithful the image will be. This random draw can be done with or without replacement. Malvern et al. (2004) discussed the strengths and weaknesses of these two sampling procedures. Sampling with replacement is usually preferred in order to estimate an author's vocabulary size. Sampling without replacement is recommended in order to estimate the LD of a given text, but both types are used in this context.

**Mean Type-token Ratio in Random Samples (MTTRRS).** MTTRRS refers to the mean number of types observed in *s* random samples of *n* tokens extracted with replacement (Malvern et al., 2004). It is convenient to divide this mean by *n* to obtain a TTR. Lu (2012) and Nasseri and Thompson (2021) set *s* to 10 and *n* to 50.

**The Probability Distribution Family: HD-D (and Vocd), Simpson's D and Yule's K.** Instead of extracting random samples, it is possible to use an exact probability calculation to determine the number of types that would be observed on average if all possible samples of a given size were extracted from a text. Baayen (2001) called this approach "interpolation". In the case of sampling without replacement, the hypergeometric distribution has to be used, and the LD index computed in this way is HD-D, as proposed by McCarthy and Jarvis (2007). The hypergeometric distribution allows for the computation of the probability that any type that is present with a given frequency in the entire text would be found at least once in a sample of size *n*. Adding these probabilities for all the types present in the original text produces the number of types that are expected in a sample of *n* tokens. Dividing this sum by *n* provides the expected TTR for a sample of *n* tokens; that is, HD-D.

McCarthy and Jarvis (2007) proposed this index in a paper that presented a theoretical and empirical analysis of the renowned index vocd described in detail in Malvern et al. (2004). They showed that vocd, which is based on the number of types observed in samples obtained via a random draw without replacement, approximated HD-D. Studies comparing these two indices have consistently reported extremely high correlations between them, which are usually well above 0.90 (McCarthy & Jarvis, 2007, 2010). McCarthy and Jarvis (2007) proposed setting the parameter *n* to 42, referring precisely to the way in which vocd is computed.

HD-D is also a generalization of the D-index proposed in ecology by Simpson (1949), but rarely employed in the study of LD (Jarvis, 2013). When applied to the lexical domain, Simpson's (1949) concentration index is equal to the probability of observing a single type when a sample of two tokens is extracted from a text without replacement. When this value is subtracted from 1, we obtain the Gini-Simpson index, which corresponds to the probability of observing two different types in a sample of two tokens; that is, HD-D with *n* set to 2.

If the binomial distribution is used, and thus sampling with replacement, the index obtained could be called Binomial Distribution Diversity (BD-D). Just as HD-D is a generalization of Simpson's D, BD-D is a generalization of Yule's *k*, since *k* approximates the probability of observing two different types in a sample of two tokens obtained with replacement (Simpson, 1949).

In summary, HD-D can be seen as the archetype in this category. The LD value it provides depends on a single parameter; that is, the arbitrary length to which each text is reduced.

## Indices Taking a Local View of LD

These indices are based on text segments; that is, sequences of contiguous tokens. Unlike the indices in the first category, except for MTTRRS, they are not defined by a mathematical formula, but by a computational algorithm. These indices are differentiated by the way they define the analyzed segments. The procedure can be characterized according to whether it is systematic, random, or data-dependent, whether the segments are independent (that is, a given token cannot be present in more than one segment), and whether it assigns the same weight to each token to calculate the score (see Appendix S2 for an explanation of how these weights are calculated).

### *Mean Type-token Ratio in Sequential Samples (MTTRSS)*

This index is equal to the average number of types observed in *s* samples of *n* contiguous tokens selected from a first token that was chosen randomly (with replacement) in the text (Malvern et al. 2004). It is convenient to divide this average by *n* to obtain a TTR. MTTRRS and MTTRSS are often mentioned together in the literature, and the same parameter values are used (*s* = 10 and *n* = 50). The segments used by MTTRSS are defined by a random process; they are (usually) non-independent, and the tokens between position 1 and position *n*-1 and those between position *N*-*n*+1 and position *N* receive a different weight, which is linearly related to the distance between their position and the closest extreme position (1 or *N*).

### *Mean Segmental TTR (MSTTR)*

This index is obtained by dividing a text into contiguous segments of *n* tokens and calculating the average TTR of each of these segments (Malvern et al., 2004). The parameter *n* is usually set to 50. This index is systematic; the segments are independent, and the index assigns a zero-one weight to the tokens except in the case that the division of *N* by *n* produces an integer.

### *Moving-Average Type-Token Ratio (MATTR)*

This index is obtained by moving a window of *n* tokens along the tokens of the text, starting with the first one and advancing by one token each time. The final score is the average of the TTRs calculated in each window (Covington & McFall, 2010). The value of the parameter *n* is usually set to 50 in research on language learning. The segments are defined in a systematic way; they are non-independent, and the tokens receive a variable weight exactly as is the case with MTTRSS.

### *Measure of Textual LD (MTLD)*

MTLD is equal to the mean length of sequential token strings in a text that maintains a TTR at a certain level, the parameter of the procedure, which is usually set at 0.72 (McCarthy & Jarvis, 2010). MTLD is obtained by starting a first segment with the first 10 tokens in the text and extending it by one token each time, unless the TTR computed in this segment is less than 0.72. When this occurs, a new segment of 10 tokens is started. The (likely) incomplete segment at the end of the text is used to estimate the number of words it would need to contain to reach a TTR of 0.72. This procedure is performed a second time, starting at the end of the text. The final MTLD value is the average length of all the segments thus formed.

Recently, Vidal and Jarvis (2020) proposed a new version of this index called MTLD-W that reduces the variability problem caused by the final segment when calculating MTLD. Since MTLD and MTLD-W produce highly correlated LD scores (Vidal & Jarvis, 2020; Kyle et al., 2021), MTLD was preferred to MTLD-W in this study due to the limited number of studies using the latter.

The length of the segments for MTLD is defined by the data; they are independent. The question of weighting does not arise in the same way that it does for the other indices. One could consider that all the tokens have the same impact on the computation procedure; that is, they increase or decrease the TTR of the current segment. However, this is not entirely correct because the first ten tokens in a segment are treated as a single block, and because the tokens in the (likely) incomplete last segment are used to estimate the length of that segment and not to compute it.

**The Twofold Length Problem and How to Evaluate It**

In the language sciences, almost all research on the validity of LD indices has focused on the sensitivity of these indices to the length of texts, a problem that is complex to address because it requires the comparison of text excerpts of different lengths but with lexical content as similar as possible. The second problem discussed in this section, namely the sensitivity to the parameter that determines the size of the samples of texts used, has received much less attention.

**The First Problem: Evaluation of Text Length Sensitivity**

The two main methods used in the literature to determine whether an LD index is sensitive to text length are the parallel sampling method and the random sampling method. Following a discussion of their advantages and limitations, two complementary methods are proposed.

*The Parallel Sampling Method*

The parallel sampling method is undoubtedly the method that is used most frequently to evaluate the LD indices. First, all texts are truncated to the same length, such as 300 tokens. The LD for this (truncated) text is then compared to the average LD of the segments obtained by dividing the text into two, three, or four; that is, the average of the two 150-word segments, the three 100-word segments, and the four 75-word segments. This method allows for the comparison of segments of different lengths, but which contain exactly the same tokens when all segments of a given length are added together. If all the values of an LD index are (almost) identical, this index is considered not to be affected by text length.

The parallel sampling method does not allow for the valid evaluation of either global or local indices. The problem is that, if all the tokens of a text are present in the sum of the segments of a given size, they do not co-occur with the same tokens. In the longest extracts, global indices such as HD-D take the repetitions of the same type at long distances into account. They cannot do so when these extracts are divided into smaller segments (Malvern et al., 2004). Consider the example presented previously: If the conclusion of an essay repeats the introduction extensively, the global approach will take the repetitions into account when analyzing the complete text, but will not be able to do so when analyzing the segments[3]. Even if this problem is less severe for local indices because they process only a subset of the tokens of a segment simultaneously, the parallel sampling method is also not suitable to evaluate them. For example, due to moving the window along the segment, MATTR does not have access to the same information in a segment of a given size and in two segments obtained by dividing it into two.

*The Random Sampling Method*

A second method for evaluating the length sensitivity of the indices is to randomly permutate all the tokens in a text and to then calculate the LD of the first $m$ tokens in this random sequence[4] (Tweedie & Baayen, 1998). An insensitive index should always produce an extremely similar value regardless of the value of $m$. A large number of permutations is necessary to reduce the impact of the random variability. This method solves the problem of the parallel sampling method because all the tokens have the same likelihood of being present with any other token in a segment of a given size.

This random sampling method can only confirm the stability of the global indices based on random sampling without replacement for the simple reason that the random permutations approximate the hypergeometric distribution used by these indices. This method is applicable to indices based on the transformation of the number of types and tokens. The situation is more complex for local indices. On one hand, they can be evaluated using this method if one considers that the order in which the tokens are extracted corresponds to the sequential order of the sample in question. On the other hand, this method denies "the non-random way in which words are used in actual coherent prose" (Tweedie & Baayen, 1998, p. 332).

### The Ordered Random Sampling Method

It is easy to make the random sampling method more compatible with local indices, as it is sufficient to reorder the randomly extracted sample of tokens according to the order of these tokens in the original text; this means that the indices are applied to a gap-fill text in which a number of random tokens have been deleted. This method is not necessary for the global indices because they are insensitive to the order of the tokens in the sample, but it allows for a better understanding of the impact of sequentiality on local indices.

### The Alternating Token Sampling Method

A limitation of the ordered random sampling method is that the extracted tokens in a given sample may not be distributed in a relatively homogeneous way across the entire text. This drawback can be overcome by adapting the split-half procedure proposed by McKee et al. (2000), which consists of comparing the LD of the entire text to the average of those obtained for even-numbered or odd-numbered tokens. It is easy to generalize this method to produce several sample sizes by taking one token out of three and one out of four. It is necessary to truncate the texts to the same length in order to guarantee that the samples in a given condition have precisely the same size in all texts.

A weakness of the above-mentioned method is that the decision to place tokens 1, 3, 5, 7... in the same sample is arbitrary; it would be as justifiable to place tokens 1, 4, 5, 8... together. The impact of the chosen division can be reduced by randomly sampling a large number of these sequences. To obtain the 1-in-4 token samples for instance, the text is divided into segments of four consecutive tokens starting from the first one and a random procedure is used to independently distribute one token of the four in each sample. This version of the method, which is used in the following, also has the advantage of being able to select two tokens in the same sample that follow each other in the original text.

## The Second Problem: The Impact of the Index Parameter

The second potential problem with the LD indices concerns indices that use samples or segments to evaluate LD; that is, all the indices discussed except TTR and the transformations. These indices use a parameter that directly (or indirectly in the case of MTLD) determines the number of tokens that will be used to evaluate LD. This parameter is systematically set in an arbitrary way or on the basis of preliminary analyses. To my knowledge, its impact on the conclusions of the analyses is almost never considered in studies of language learning.

However, Covington and McFall (2010, p. 97) underlined the impact of this parameter in MATTR: "Obviously, the moving-average TTR of a text varies with the window size more or less the same way that the conventional TTR varies with the text length. " The authors indicated that the window size should be determined according to the objectives of the study. They also pointed out that MATTR will be more sensitive to short- or long-term repetitions depending on the window size. If two texts differ in the proximity of repetitions, the choice of the window size will affect the results.

With regard to HD-D, work in ecology on Simpson's index, a special case of HD-D, has shown that it is only the most extreme case of a family of indices obtained by varying the size to which samples are reduced, thus the parameter. Hurlbert (1971, p. 581) emphasized the role of this parameter in these terms:

Species richness comparisons made at a single sample size (n) permit only limited conclusions. Since the manner in which sample species richness increases with sample size varies according to the number of species and their relative abundances in the collection, it is possible that at one sample size, collection A will have a greater sample species richness than collection B, while at a larger sample size, collection B will have the greater sample species richness.

Determining the impact of this parameter in the language learning field is the second objective of this study.

## Method

### Data Collections

Two datasets consisting of written essays that were previously employed in LD studies in Language Learning (Bestgen, 2017; Zenker & Kyle, 2021) were used to evaluate the sensitivity of the indices to text length. The texts in the two datasets differ in length; one contains texts of 400 words or less, while the other contains texts of 480 to 895 words, thus allowing for analyses to be conducted over a wide range of text lengths. Since these texts were rated for quality, they can be used to determine whether length effects are sufficient to significantly alter the findings of studies linking LD and text quality, which is probably the most frequently addressed question in this field.

A limitation of these datasets is that they only represent a single type of learner production data; that is, written essays. To make the results of this study more generalizable, a third dataset consisting of English learner monologues from the Corpus of English as a Foreign Language was analyzed. The conclusions reached in these analyses are identical to those reported below. Due to a lack of space, these analyses are presented in Appendix S6.

### *International Corpus Network of Asian Learners of English (ICNALE)*

The ICNALE Edited Essays (V2.1, 2018 July) is described in Ishikawa (2018). It is a subset of the ICNALE written corpus, and consists of 640 argumentative essays of between 180 and 400 words in length that were written by 320 EFL and ESL learners from 10 Asian countries. Compared to the much larger parent corpus, it has the advantage of the texts having been evaluated by professional editors using the five rating rubrics in the ESL Composition Profile (Jacobs et al., 1981): content, organization, vocabulary, language use, and mechanics. The measure of quality used in the following is the weighted average of the scores obtained using the five rubrics. A limitation of these evaluations is that each essay was judged by only one rater, but a small-scale calibration study was performed in which the five raters were asked to grade the same set of eight essays. The average correlation for all the raters taken two by two was .84 and the minimum correlation was .74, which is an acceptable level of interrater reliability for this type of task (Artstein & Poesio, 2008).

### *International Corpus of Learner English (ICLE)*

This dataset is described in Thewissen (2013), and includes 223 argumentative essays of between 450 and 900 words in length that were written by French, German, and Spanish learners from the ICLEv1. Two professional raters were asked to rate each text according to five rubrics of the Common European Framework of Reference Writing Scale; that is, vocabulary accuracy, grammatical accuracy, orthographic control, vocabulary range and coherence/cohesion. They were then asked to allocate a holistic score. Each essay was rated as being B1, B2, C1, or C2. In addition, the raters were allowed to use + or - signs to further differentiate the sub-levels. These ratings were transformed into a numerical scale. Holistic ratings, which were highly correlated with the score for each rubric (r > .95), were used in the following analyses. The Pearson correlation coefficient between the values given by the two raters was .69, which corresponds to a Spearman-Brown reliability of .82.

## Procedure
### Text Preprocessing
CLAWS7 (Rayson, 2003) was used to extract all the tokens; that is, words in the sense of orthographic forms. As in Baayen (2001), no lemmatization was applied, tokens such as *be*, *is* and *were* being treated as different types. In order to be able to truncate the texts to a sufficient length for the analyses, a minimum length of 240 tokens was imposed in ICNALE, thus reducing the number of texts in this dataset to 188. No texts were deleted in ICLE because they all contained more than 400 tokens. Table 1 presents descriptive statititics for the two datasets.

**Table 1**
*Descriptive Statistics for the ICNALE and ICLE Datasets*

| Dataset | n | Text length (in tokens) | | | |
|---|---|---|---|---|---|
| | | Mean | Std | Min | Max |
| ICNALE | 188 | 272 | 26 | 240 | 400 |
| ICLE | 223 | 681 | 95 | 485 | 890 |

### Evaluated Indices
The following ten indices were compared: the TTR, as a base level, and three transformations, Guiraud's R, Herdan's C, and Maas's a, two global indices, MTTRRS and HD-D, and four local indices, MATTR, MSTTR, MTTRSS and MTLD.

## The First Problem: *Evaluation of Text Length Sensitivity*
For these analyses, the parameter *n* was set to 50 and *s* to 10. For MTLD, the TTR factor was set to 0.72. All the texts in each dataset were reduced to the same length, namely 240 tokens for ICNALE and 400 for ICLE. This is the usual way to proceed in the parallel sampling method because it guarantees that the segments obtained after the division have the same length in all the texts that are compared. It is also necessary for the alternating token sampling method, but is not necessary for the two random sampling methods. Nevertheless, it was considered preferable to also truncate the texts for these methods in order to place all the methods under the same conditions.

The parallel sampling method was applied to each truncated text by dividing it by 2, 3 and 4, and the LD indices were calculated for the complete extract and for all the segments. The data for the statistical analyses were the index values for the complete extract and the average of the values for the segments of the same length. The same lengths as those produced by the parallel sampling method were used for the random sampling method; that is, 240, 120, 80, and 60 for ICNALE, and 400, 200, 133, and 100 for ICLE. A single score is obtained when the entire truncated text is analyzed. Ten thousand random permutations were performed in the other three conditions, and the final score for a condition was the average of the scores obtained for these permutations. The text samples that were analyzed using the ordered random sampling method were identical to those that were analyzed using the random sampling approach; the only difference was that the tokens were replaced in their original order before the indices were computed. Finally, for the alternating token sampling method, four lengths were produced by selecting all the tokens or one token out of two, three or four. Ten thousand random selections were made for these last three conditions, and the final score was the average of these scores. To facilitate the reproduction of the study, Appendix S9 provides python scripts that implement the least often used methods.

**The Second Problem:** *Manipulation of the Index Parameter*

The texts were not truncated for this analysis, since the objective was to mimic a standard LD study in which texts of different lengths are compared. This analysis could not be performed on the TTR and the three transformations because these indices do not use a parameter. In order to set the values of the parameters to be evaluated, I relied on the factor size analysis of MTLD presented by McCarthy and Jarvis (2010), who indicated that values ranging from 0.66 to 0.75 could be used; that is, ten different values when rounded to the second decimal place. Ten values were also used for the other indices in which the parameter determined the sample size. For ICNALE, the selected values ranged from 24 to 240 tokens in steps of 24 and from 40 to 400 in steps of 40 for ICLE. This way of setting the parameter value probably favored MTLD because the values proposed by McCarthy and Jarvis (2010) were not expected to affect this index.

**Statistical Analysis**
*Critique of the Usual Statistical Technique*

The evaluation methods all generated the same type of data to evaluate the impact of length, namely an LD score computed for each sample length that was extracted from each analyzed text. When the analyses are performed on a single text, the results are presented in the form of a graph showing the evolution of the scores according to the sample length. An index is considered to be unaffected by length if the LD score is constant for all lengths.

However, in the field of language learning, several tens or hundreds of texts are usually analyzed, thus making the use of graphics difficult. The effect of length is then almost always (but see Fergadiotis et al. (2013, 2015) for indirect approaches based on confirmatory factor analysis and structural equation modeling) assessed by means of an ANOVA with a repeated factor. The underlying assumption is that, if the sample length has no impact, the means of the LD scores for each length should be approximately the same, and the ANOVA should therefore be statistically non-significant.

This approach has two important flaws, the most obvious being that it distorts the correct use of null-hypothesis significance tests, which are designed to reject the null hypothesis (Norris, 2015). The absence of a statistically significant difference does not validate an index. The second flaw is that a comparison of means employed in this context is only aimed at detecting systematic biases between different sample lengths. However, an index can be problematic even though the mean values per length are not statistically significantly different. It is sufficient that the impact of the length is different for each text, thus producing a text × length error that overrides the overall effect of length (Bruton et al., 2000).

**Table 2**

*Means for the Four Lengths and ANOVA Results for the Parallel and Random Sampling Methods*

| Sampling Method | Sample length | | | | $F$ | $p$ | Partial *eta*$^2$ |
|---|---|---|---|---|---|---|---|
| | 60 | 80 | 120 | 240 | | | |
| Parallel | .7740 | .7744 | .7750 | .7753 | 0.44 | .67 | .007 |
| Random | .7803 | .7803 | .7802 | .7802 | 3.83 | .03 | .058 |

*Note*: $N = 188$, Dfs = 3, 187. Appendix S3 provides additional information.

These flaws can be illustrated by the following example taken from the analyses presented in the Results section. The parallel sampling method and the random sampling method were applied to MATTR in the ICNALE corpus. Table 2 shows the means for the four sample lengths and the results of the repeated measures ANOVA. In these analyses, no violations of the underlying assumptions of normality and sphericity were observed.

**Figure 1**

*Profile Graphs for MATTR in ICNALE by the Parallel Sampling (Left) and Random Sampling (Right) Methods*

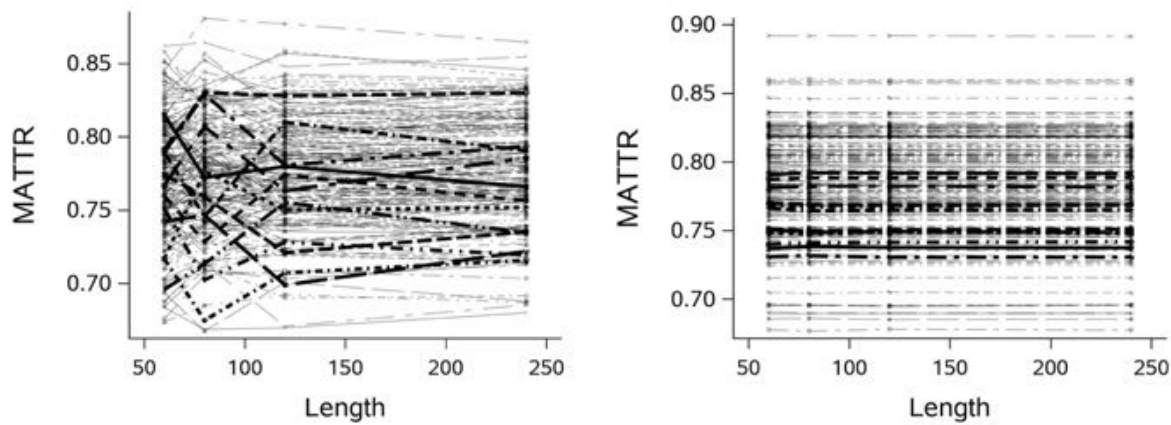

The table shows that the means were relatively constant. The ones that evolved the most were obtained via the parallel sampling method, the *p*-value of which nevertheless pleads for non-significance, using an alpha of .05, in contrast to that of the random sampling method. Figure 1 (see below for detailed explanations regarding how these figures were drawn) explains these observations. According to the random sampling method, the MATTR scores for each text were almost constant; so constant that an extremely small decrease in the mean was detected by the F-test. The profiles were completely different when the parallel sampling method was used. Large variations in the LD score were observed for many texts, thus producing crossovers between lines. These crossovers imply that a text that is considered to be less diverse than another when it is divided into segments of 60 tokens is much more diverse when it is divided into segments of 120 tokens. These profiles suggest an impact of length on MATTR when the parallel sampling method is used, whereas the random sampling method shows almost no sensitivity, which is the opposite of the conclusions drawn from the ANOVA. These results also show that a significant ANOVA does not necessarily imply an important sensitivity of an index to the text length.

### *A Better Approach: The Intraclass Correlation Coefficient*

For an index to be considered to be completely unaffected by text length, it is necessary for any differences in the scores for samples of the same text of different lengths to only be due to errors that are related to the method used, such as the impact of the random permutations. The intraclass correlation coefficient (ICC) is the traditional procedure for evaluating this condition using interval data (Tinsley & Weiss, 1975). It corresponds to the proportion of the total variance in the scores due to variance in the rated object (in this case, the text), and is therefore a measure of the effect size. The closer it is to 1, the less the index is affected by the sample lengths. ICCs are based on two assumptions: the independence of the raters and the ratees, which must be guaranteed by the study design, and the normality of ratings. A major advantage of the ICC is that, even when the data are non-normal, the estimation of the appropriate components of variance and the reliability coefficients are still permitted (Gildera et al., 2007; Landis & Koch, 1977).

There are several ICCs depending on the design and objectives of the study (Shrout & Fleiss, 1979). In the current case, ICC(2,1) has to be used because the different lengths that were compared can be seen as a (relatively) random sample of the set of lengths that could be evaluated, and each of these lengths was applied to each of the texts that were analyzed. McGraw and Wong (1996) presented two versions of ICC(2,1) depending on whether the objective was to measure absolute agreement, for which the scores for the different sample lengths of a text must be identical, or the consistency, for which it is sufficient for the scores to be proportional when expressed as deviations from their means; that is, correlated. Absolute agreement was the goal in the present case, except when the analyses concerned the impact of the index parameter, as it was expected that this parameter would modify the mean LD scores. However, it was hoped that this manipulation would not alter the relative positions of the texts with regard to their LD level.

The minimum acceptable level of an ICC depends on the goals of the study. As Nunnally (1978, as cited in Lance et al., 2006, p. 206) pointed out, "In those applied settings where important decisions are made with respect to specific test scores, a reliability of .90 is the minimum that should be tolerated, and a reliability of .95 should be considered the desirable standard. " As the LD indices were intended to cancel out any differences related to the length of the texts, ICCs as close to 1 as possible were desirable. However, the fact that error variance may also be due to the method that was used in the evaluation should also be kept in mind.

### A Complementary Approach: Profile Graphs

Although the ICC has the advantage of being an objective measure of effect size, it does not allow for the visualization of the LD text profiles. Thus, it was complemented in the results by graphs of the same type as those used when the analysis concerned only one text. In order to make these graphs as informative as possible, the twelve profiles that had the most differences that were extremely large between conditions taken two by two, will be made more visible. The same criterion and the same procedure were applied to all the indices in order not to favor any of them. The procedure that was used is explained in Appendix S4.

### Software used

All of the statistical analyses were performed using SAS software, except the calculation of the ICC confidence intervals which was performed with the IRR package (R Core Team, 2021). The SAS code and the R function used can be found in Appendix S8. Appendix S7 explains how to obtain the main LD indices analyzed in this paper by means of two freely available tools, TAALED (Kyle et al., 2021) and the koRpus package for R (Michalke, 2020).

## Results and Discussion

### The First Problem: Evaluation of the Text Length Sensitivity

#### ICC Analysis

Figure 2 shows the ICCs for the two datasets, the four evaluation methods, and the ten indices that were analyzed. The mean ICC is represented by a circle, while the bars represent the 95% confidence interval for the ICC population values (McGraw & Wong, 1996). The graphs clearly show that the most important differences were related to the indices, with the other two manipulated factors, namely the analyzed dataset and the evaluation method, having much less impact.

All the graphs show the same division of the indices into two groups: TTR and the transformations, as well as MTTRRS, performed extremely poorly. The other five indices were far superior in terms of controlling for the impact of length because they effectively reduced all the texts to the same length using a probabilistic or an algorithmic approach. The mean ICCs were almost always above .90, and the lower bound of the confidence interval

**Figure 2**

*ICCs for Text Length Sensitivity for ICNALE (Left) and ICLE (Right)*

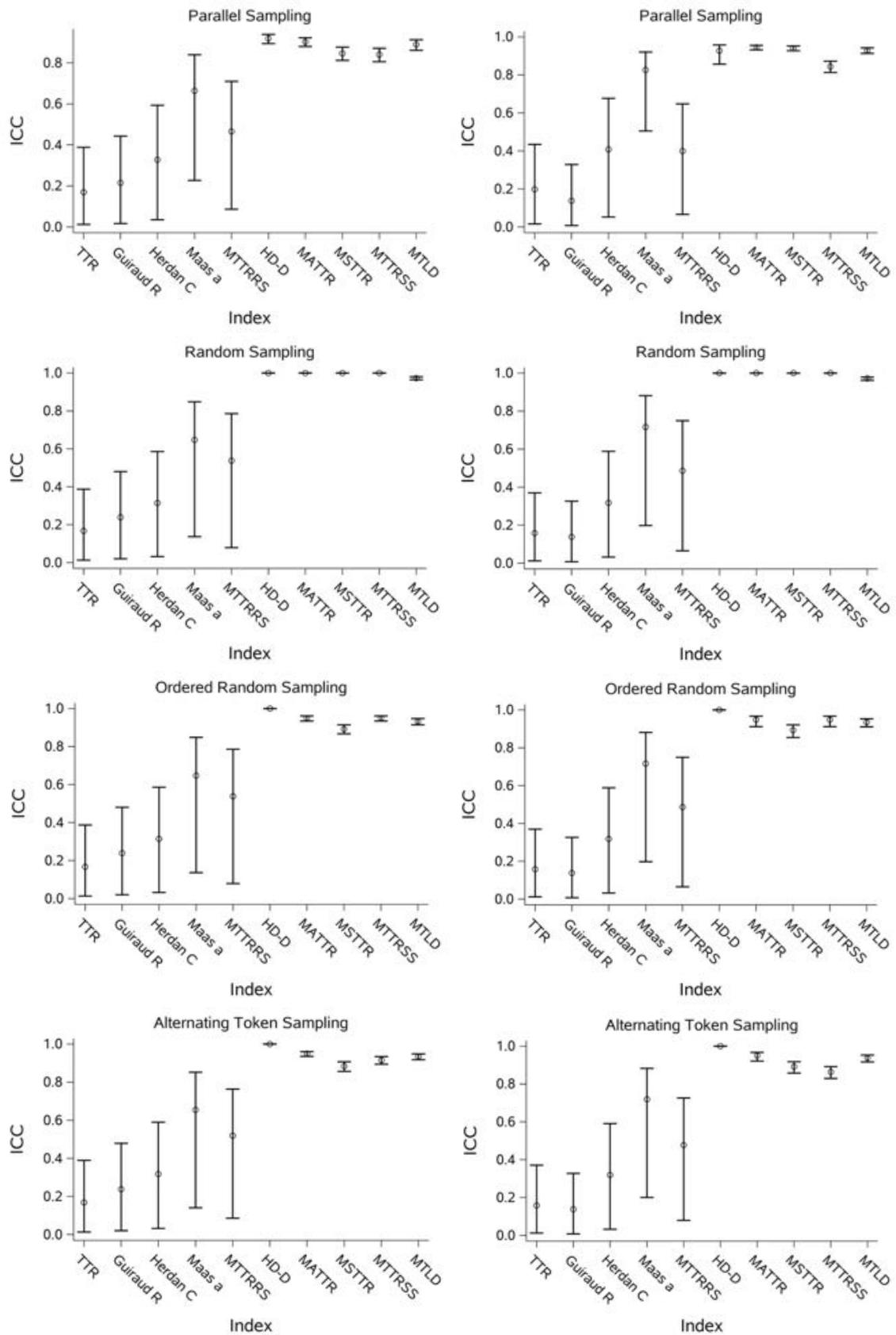

**Figure 3**

*Profile Graphs for ICNALE in the Alternating Token Sampling Method*

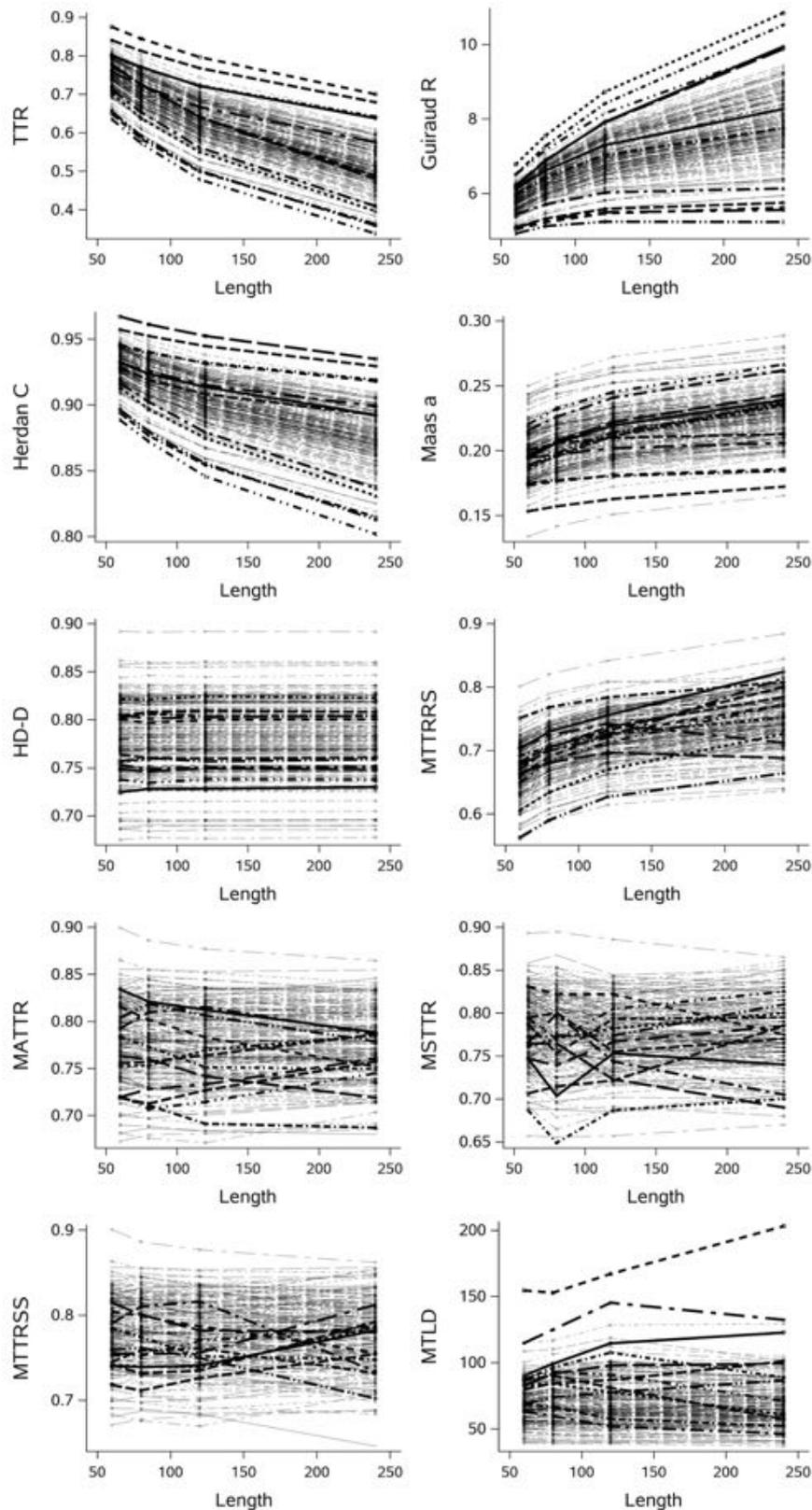

was consistently above .79. In six out of eight graphs, the ICC for HD-D was nearly equal to 1, since the smallest of these values was .9993. The differences among the other four indices varied by condition, and the confidence intervals frequently overlapped. Nevertheless, it appears that MATTR was slightly more effective than the others, followed by MTLD.

The differences among the four assessment methods deserve specific comments. As expected, the parallel sampling method penalized HD-D. A comparison of the two randomization-based methods showed that token reordering penalized MATTR, MSTTR, and MTTRSS more than it did MTLD, but that MTLD performed worse than the others with unordered randomization. The alternating token sampling method results were relatively similar to those obtained via the ordered random sampling method. It confirms the extremely high efficiency of HD-D for controlling the length effect. This conclusion contradicts that of several previous studies. An in-depth discussion of the most important studies is given in Appendix 10.

### *Graphical Representation of the Profiles*

Figure 3 shows the impact of length on the indices. These profiles were obtained using the alternating token sampling method for ICNALE. The figures for the other methods and for ICLE are provided in Appendix S1. The five indices that performed worst according to ICC clearly showed a length bias. The profiles for MATTR, MSTTR, MTTRSS, and MTLD did not show any bias, but there were important crossings, which explained why the ICC was sometimes quite far from 1. Only the profiles for HD-D were almost perfectly stable.

### *Discussion*

All four sampling methods have their advantages and disadvantages. The parallel sampling method is the only one that extracts samples composed of continuous sequences of tokens. Figure 2 shows that it can identify very length-sensitive indices. However, due to the problems mentioned above, it is not clear that it can make fine distinctions between indices that are more effective at controlling length. The alternating token sampling method is preferable because it ensures that samples of different sizes are truly comparable in content. It respects the order of the tokens in the text and their distribution within the text. This last point suggests that it is also preferable to the ordered random sampling method. One might think that the random sampling method is useless. This is far from obvious, as it has the advantage of breaking the order of the tokens in a text. This has an impact on local indices, such as MATTR and MTLD, which are sensitive to the fluctuation of LD in a text. By cancelling out any effect of token order, it allows these local indices to be compared independently of this factor. However, it also has the disadvantage of cancelling out the impact of differences in the weights assigned to tokens by certain indices, since all tokens have the same probability of being positioned at any point in the sequence and therefore of being underweighted, for example by MATTR, when they are at the ends.

The TTR, the three transformations, and MTTRSS were clearly unsatisfactory. If this conclusion was expected for the first four indices, it was less expected for MTTRSS, particularly since it does not apply to MTTRSS, although these two indices are often presented as variants of each other (Lu, 2012). MTTRRS differs not only from other indices based on random sampling, but also from indices based on text segments, in that the samples of 50 tokens are extracted with replacement[6]. The smaller the length of the sampled text, the more likely it is that the same token will be selected more than once, thus reducing the LD.

As the parallel sampling method is problematic, HD-D is clearly the index that is least affected by text length. If one wishes to favor a local index, MATTR appears to be slightly better than the others, particularly for the alternating token sampling method, but the differences are not very important. The major weakness of MTTRSS compared to MATTR is the number of segments it considers, which was set at 50 in this study based on standard practice, whereas MATTR employs N-n+1 segments; that is, 191 for a length of 240 tokens in ICNALE. MTLD often performs almost as well as MATTR, but its significantly poorer performance in the random sampling method warrants further investigation.

**Figure 4**
*ICCs for the Impact of the Index Parameter for ICNALE (Left) and ICLE (Right)*

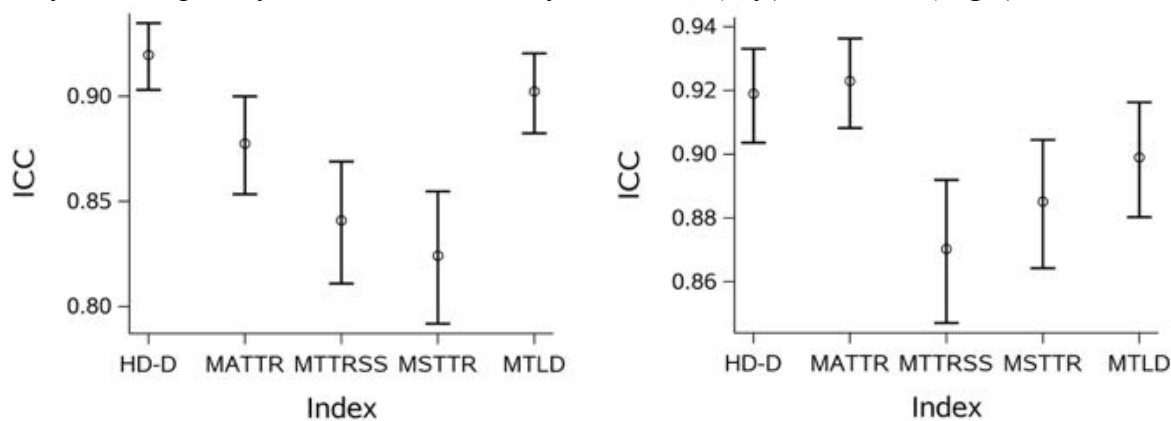

**The Second Problem: Impact of the Index Parameter**
*ICC Analysis*

Figure 4 shows the consistency (correlation) of the indices as a function of the parameter value measured by the ICC. The impact of the parameters was not negligible, since the best ICCs barely exceeded .90. HD-D was the least affected overall, followed by MATTR and MTLD.

*Graphical Representation of the Profiles*

Figure 5 illustrates the impact of parameter manipulation on the LD scores for the texts in the two datasets. In this figure, the scores for each index for each value of the parameter are centered on 0 in order to eliminate bias and to match the data used for the ICC exactly when it measures consistency. For all the indices, the LD score of some texts increased with the increase of the parameter value, while this was the opposite for other texts, and produced profile crossings. These crossings are obviously problematic since they mean that some texts, which are more lexically diverse than others with a low value of the parameter, are less diverse than are those with a high value.

HD-D and MATTR presented relatively simple profiles in the sense that they could be approximated by a low-degree polynomial curve, while the profiles of MSTTR and MTTRSS were much more erratic. For MTLD, it appeared that the profiles that were most affected are the ones for which the segments are the longest. These are the most diverse texts, but also the ones for which the last factor has the most impact. This is because the final factor, almost always incomplete, is frequently misestimated (Vidal & Jarvis, 2020). This factor has a greater impact when texts are lexically more diverse because fewer segments are obtained.

*Impact on the Correlation between LD and Text Quality*

An important question concerns the extent to which these differences due to the parameter can affect the conclusions of studies of language learning. It is difficult to provide a definitive answer to this question, but it is possible to assess whether this impact is sufficient to alter the conclusions of a traditional analysis, such as the correlation between LD scores and the text quality (Crossley et al., 2010; Lu, 2012). The correlation for each parameter value is presented in Tables 3 and 4; the largest value for each index is in bold, and the smallest is underlined. To check the underlying assumptions, the data distributions were examined graphically, as recommended by Hu and Plonsky (2021). There was no indication of non-linear relationships, nor of extreme values. The only major deviations from normality were observed for MTLD in the ICNALE and ICLE datasets (but not in the COREFL dataset). It is worth remembering that MTLD is the only index in this study that does not output a sort of TTR, but the mean number of tokens that maintains a TTR at .72. The log $e$ transformation proved effective in normalizing these distributions. The correlations obtained after this transformation were very close to those observed without it, and no conclusions were altered. The results obtained with untransformed values are reported, as this is how the MTLD score is systematically analyzed.

**Figure 5**

*Profile Graphs for the Parameter Impact for ICNALE (Left) and ICLE (Right)*

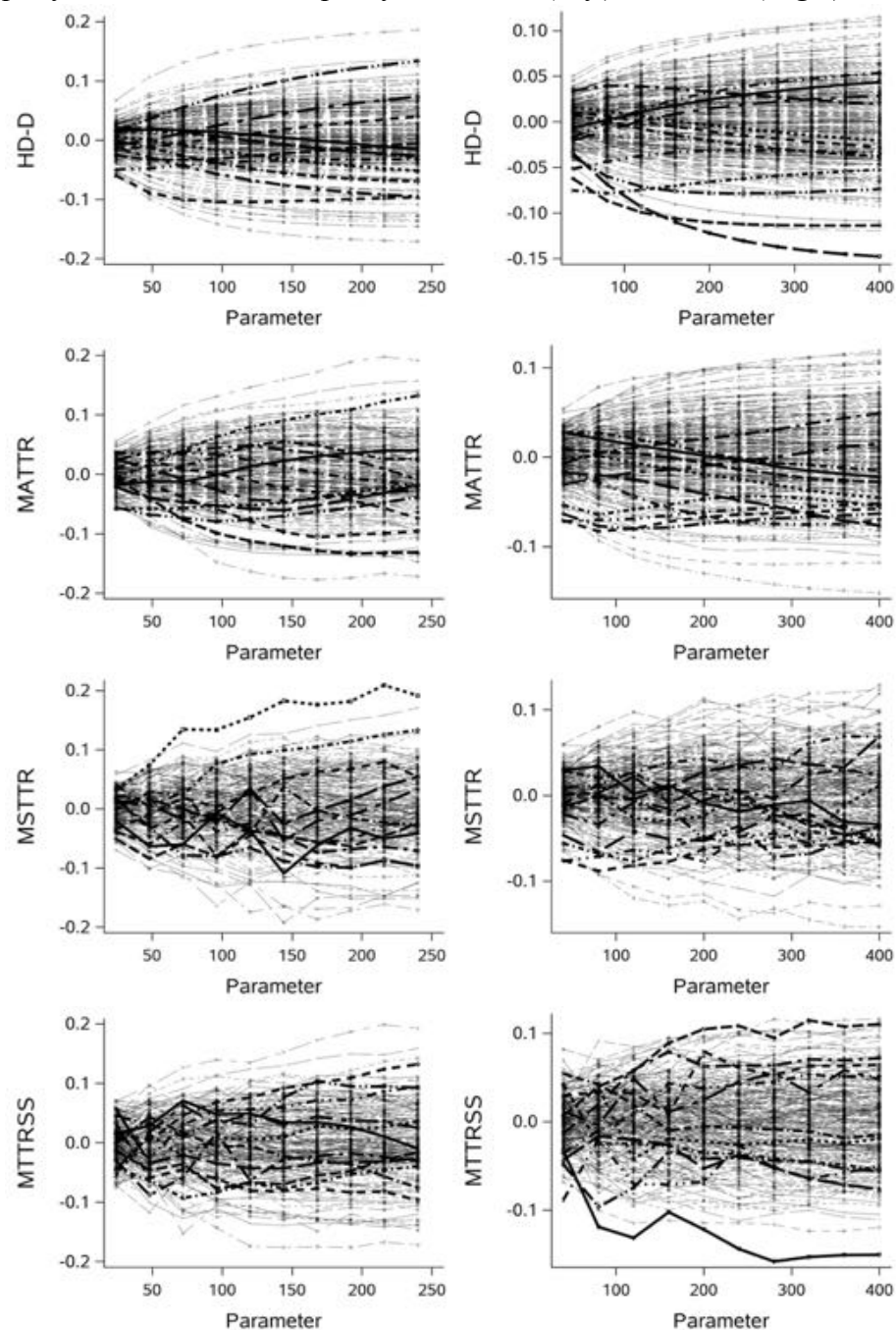

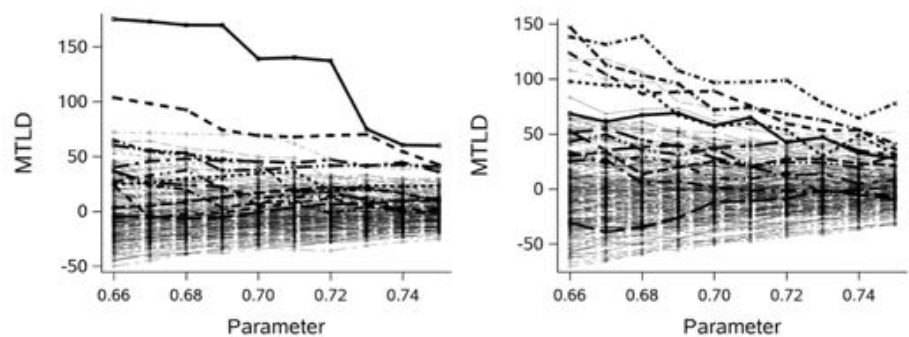

**Table 3**

*Correlations Between LD and Text Quality for ICNALE for Different Parameter Values*

| Length | HD-D | MATTR | MSTTR | MTTRSS | Factor | MTLD |
|--------|------|-------|-------|--------|--------|------|
| 24 | .285 | **.390** | **.385** | .299 | .66 | .285 |
| 48 | .306 | .367 | .331 | .371 | .67 | .289 |
| 72 | .319 | .367 | .365 | .326 | .68 | .287 |
| 96 | .327 | .370 | .344 | **.355** | .69 | .252 |
| 120 | .333 | .368 | .336 | .336 | .70 | .242 |
| 144 | .337 | .365 | .321 | .373 | .71 | .296 |
| 168 | .339 | .363 | .324 | .367 | .72 | .284 |
| 192 | .341 | .360 | .348 | .363 | .73 | .308 |
| 216 | **.342** | .347 | .340 | .340 | .74 | .307 |
| 240 | .341 | .341 | .336 | .343 | .75 | **.320** |

These tables show an impact of the parameter on the correlation, but it was somewhat limited. It was also observed that it was not always the same parameter values that produced the highest or lowest correlations in the two datasets. Steiger's (1980) test for two non-independent correlations was used to determine if there was a statistically significant difference between the largest and smallest correlation of each index with the text quality. Only two differences were statistically significant at an alpha of 0.05: MTLD in ICNALE (Difference = .078, $t(185)$ = 3.18, $p$ = .002, CI = [.029, .127]) and MTTRSS in ICLE (Difference = .177, $t(220)$ = 3.35, $p$ < .001, CI = [.072, .282]). The lower limit of the confidence interval for the difference, which was obtained using Zou's (2007) procedure, was close to 0 for MTLD. It was higher for MTTRSS, but the analysis of the ICCs (Figure 4) suggested that this index could not be recommended. The full results of this analysis are provided in Appendix S5.

**Table 4**

*Correlations Between LD and Text Quality for ICLE for Different Parameter Values*

| Length | HD-D | MATTR | MSTTR | MTTRSS | Factor | MTLD |
|--------|------|-------|-------|--------|--------|------|
| 40 | .480 | .445 | .417 | .294 | .66 | .452 |
| 80 | .510 | **.474** | .462 | .414 | .67 | **.461** |
| 120 | **.515** | .473 | .480 | .460 | .68 | .449 |
| 160 | .513 | .468 | .477 | .458 | .69 | .449 |
| 200 | .513 | .462 | .473 | .453 | .70 | .458 |
| 240 | .512 | .460 | .481 | .448 | .71 | .440 |
| 280 | .511 | .461 | .496 | .458 | .72 | .450 |
| 320 | .511 | .462 | **.501** | .465 | .73 | .446 |
| 360 | .511 | .465 | .497 | .459 | .74 | .455 |
| 400 | .511 | .471 | .482 | **.471** | .75 | .436 |

*Discussion*

None of the five indices tested was insensitive to a change in parameter. This was also the case for HD-D, which was found to be virtually immune to the effects of length in the previous analyses. How can this sensitivity be explained? For MATTR, Covington and McFall (2010) pointed out that the parameter that sets the window size determines the weight of the repetitions in the short and long terms. A small window size, even of ten tokens, allows for the analysis of very short-term repetitions, which are typical of dysfluencies, whereas a large window size takes long-distance repetitions as well as closer ones into account. It follows that, when analyzing two texts that differ in the distance between repetitions, using different window lengths will produce different MATTR scores. This explanation obviously applies to MSTTR and MTTRSS, in which the parameter performs the same function. It also applies to MTLD, albeit in a less direct way, since the TTR factor determines the length of the window that is needed to achieve that TTR. Given the well-established observation that the longer an extract is, the more its TTR decreases, requiring a high TTR tends to produce small extracts, while a low TTR usually requires longer samples.

For HD-D, the explanation has been known in ecology since the 1970s. In this field, HD-D is part of the family of the generalized Simpson's measures proposed by Hurlbert (1971). It is a family of diversity indices precisely because of the parameter *n*. As Smith and Grassle (1977, p. 284) stated, "For large [n], the measure is sensitive to the rare species in the population while for small [n], the measure is dominated by the abundant species. "

The reason that a high value of the parameter is necessary in order for rare words to have a significant impact on LD is presented graphically in Figure 6. It shows the probability that types with a frequency between one and 20 in a text of 300 tokens will be present in a sample with a size varying from ten to 300 tokens. When divided by the sample size, this probability is the participation of this type in HD-D. As can be seen, a type that is only present once has a probability that increases linearly with the length of the extract. The probability of a type being present 20 times increases much faster at the beginning to approach its maximum of 1, even for small sample sizes. As soon as the sample contains about 50 tokens, the probability that this type will be present in the sample barely increases. The more frequent a type is, the closer its behavior is to the one that is present 20 times; therefore, it increases the LD of small samples more than it increases the LD of large ones. The opposite is obviously true the rarer a type is. This explanation of the impact of the parameter *n* on HD-D allows us to understand the counter-intuitive observation by McCarthy and Jarvis (2007), namely that the addition of an extra exemplar of a type that appears only once in a text can lead to an increase in HD-D and thus in LD. This result is due to the increased likelihood that a very rare type will be present in a sample with a limited size.

## Conclusion

The impact of text length on LD estimation is a problem that has resisted the efforts of the scientific community for more than a century. The methodological and empirical analyses presented here, based on work in both quantitative linguistics and ecology, show that the problem is twofold. The solution to the first problem, which is the dependency on text length that has received almost exclusive attention in linguistics, has in fact been known and implemented in indices for many years. It consists of reducing all texts to the same length using a probabilistic approach or via an algorithmic procedure. When evaluated using appropriate methods, these indices are not affected or are barely affected by different text lengths.

**Figure 6**

*Probability of the Occurrences of Types with Various Frequencies in Samples of Increasing Sizes in a Text of 300 Tokens*

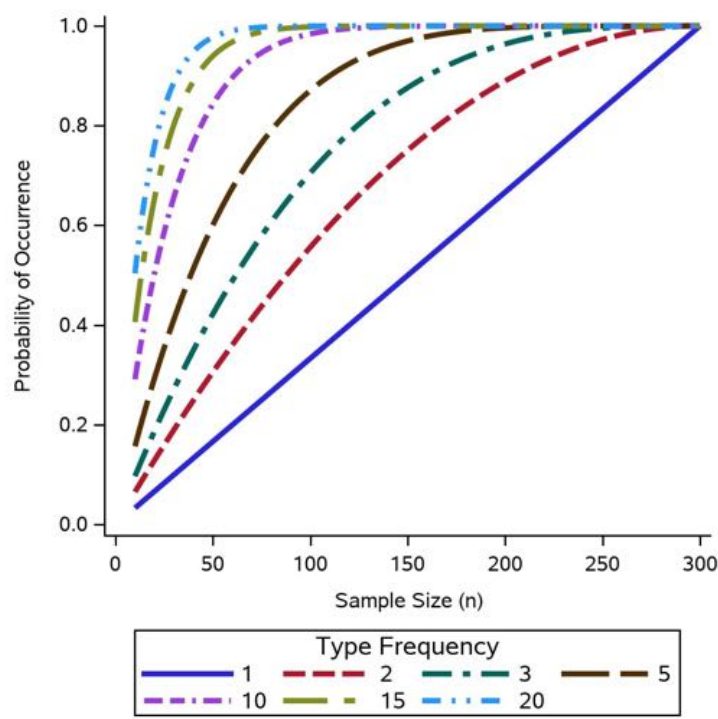

However, these successful indices present a second problem, namely their sensitivity to the parameter that determines the length of the segments to which the texts are reduced. All of them are sensitive to this, and the explanations for this effect suggest that no index based on such an approach will be insensitive. This sensitivity means that some texts, which are more lexically diverse than others with a low value of the parameter, are less diverse than are those with a high value. Analyses linking LD and text quality, performed in three datasets, fortunately suggest that this problem may only have a limited impact on the conclusions of a study. Nevertheless, further studies are necessary.

Further studies would also be useful to complement the four sampling-based methods that have been used to evaluate text-length sensitivity. As the compared samples do not form a real text in any of these methods, it may be interesting to develop approaches in which real texts with content that is as similar as possible, but which have different lengths, can be compared. As suggested by a reviewer, this could be done by requiring learners to write complete essays of different lengths on the same topic. A potential weakness of this approach is that the production situation could be seen as unnatural due to the length requirement, which might imply extensive post-editing of the text by the author. However, this situation is not particularly different from that of the author of a manuscript who is required to write an abstract of 150 words and an accessible summary of 600 words.

This methodological review leads to several recommendations to make the study of LD more valid. First, all the LD indices evaluated here, which do not reduce texts to the same length by some type of subsampling, are to be avoided because they lead to the confusion of two factors, LD and text length, particularly when this length is potentially related to the phenomenon being investigated, as is frequently the case when studying language acquisition or impairment. This is especially true of the Guiraud index, which is still frequently used in studies in language learning.

Of the global indices, HD-D is recommended. The question of the parameter value then arises. A first option, which is more pragmatic than conceptually justified, is to calculate HD-D using several parameter values and to ensure that the conclusions of the study are not profoundly modified by any of them. Other recommendations are hopefully possible. If one assumes that rare words are at least as relevant as frequent words for estimating LD, thus assuming that the author uses them correctly and that there is no spelling error, the parameter should be set to the largest possible value (see Figure 6); that is, the length of the shortest text to be analyzed. The weakness of this proposal is that it encourages researchers to use different parameter values for each study, making inter-study comparisons difficult. However, it is rare to compare the raw LD scores obtained in a study with those from other studies because many other factors influence these raw scores such as the writing prompts used (Zenker & Kyle, 2021). Similarly, in a meta-analysis, this parameter will be only one of many differences among the studies that are being compared. If this issue of comparability is considered to be the most important, setting the HD-D parameter to 42 should be preferred. This value is sufficiently low to be applied to the minimum text lengths that are usually studied (Koizumi & In'nami, 2012). Moreover, it has been used previously, and is compatible with vocd (McCarthy & Jarvis, 2007); therefore, it is also compatible with the numerous studies that have used this index.

If local estimation of LD is desired, MATTR and MTLD can be recommended. MATTR is less sensitive to length than is MTTRSS and MSTTR. With regard to the value of the parameter on which MATTR is based, the situation is different from that of HD-D. In the case of MATTR, this parameter regulates the knowledge span of the procedure and therefore its sensitivity to short-, medium-, and long-term repetitions. This parameter must therefore depend on the objectives of the study (Covington and McFall, 2010). If further studies show that it only affects the conclusions of the analyses very moderately, setting it at 50, as is frequently the case, would maximize comparability with previous studies.

MTLD appeared to be slightly more sensitive to length than MATTR, particularly in the random sampling method, but the differences were insufficient to declare one superior to the other. Nevertheless, the presence of some extreme profiles (see Figure 3) suggested that further analyses of MTLD would be quite useful, particularly since new versions of this index have recently been proposed (Vidal & Jarvis, 2021). These analyses would also be helpful for proposing guidelines for the parameter.

In summary, this study has resulted in mixed conclusions. On one hand, it shows that the problem of length that has attracted significant attention in linguistics was solved a long time ago, provided that an adequate methodology was used to evaluate it. On the other hand, a length problem remains. Fortunately, its origin can be explained, and it is therefore possible to take it into account in the analyses by fixing the parameter according to the objectives of the study. Setting the value of a parameter for an analysis is a normal and frequent practice in quantitative research. For example, it occurs in linguistics when a threshold must be set to differentiate the analyzed cases from others. In the present case, some guidelines are proposed above, but further research is necessary as this second length problem remains almost unconsidered in studies of language learning.

**Notes**

[1] These indices are used in many of the works mentioned in the references; they are not systematically recalled in the following sections.

[2] The term "local" is preferred to "sequential" as the name for this category not only because it insists on the limited memory of these indices, but also because the indices in this category are usually presented in connection with the Mean Segmental TTR (Covington & McFall,

2010; McCarthy & Jarvis, 2010), while the order of the tokens within the segments on which it is based have no impact.

3 Taking long-range repetitions in the longer extracts of the parallel sampling method into account does not imply that HD-D will always be lower the longer the extract is. In fact, as shown by McCarthy and Jarvis (2007) and as explained in the discussion of Figure 6, the presence of two tokens of the same type can increase LD.

4 The symbol $m$ is used here to distinguish the length of a sample in this evaluation method from the length of a segment or a sample used to calculate an index ($n$, the parameter).

6 MTTRSS also uses replacement sampling, as the first token in a sample can be selected multiple times, but a given token cannot be selected twice in a sample.

**Supporting Information**

Additional Supporting Information may be found in the online version of this article at the
publisher's website and below:

Appendix S1. **Profile Graphs for the Four Methods and the two data collections, except
for those shown in Figure 3**

Appendix S2. **Weight of Each Token in an LD Index.**

Appendix S3. **Additional Information for Table 2.**

Appendix S4. **Procedure for Selecting 12 Profiles in the Graphs.**

Appendix S5. **Results of Steiger's (1980) Test.**

Appendix S6. **Analysis of English Learners' Monologues.**

Appendix S7. **Obtaining the Main LD Indices by Means of Two Freely Available Tools.**

Appendix S8. **SAS Code and R Function Used.**

Appendix S9. **Implementing the Sampling Methods in Python.**

Appendix 10. **In-depth Discussion of the Most Important Studies Claiming that HD-D is
Sensitive to Text Length.**

**Supporting Information for: Bestgen Y. Measuring Lexical Diversity in Texts: The Twofold Length Problem.**

Appendix S1: Profile Graphs for the Four Methods and the Two Data Collections, Except for Those shown in Figure 3

**Figure S1.1**
*Profile Graphs for ICNALE in the Parallel Sampling Method*

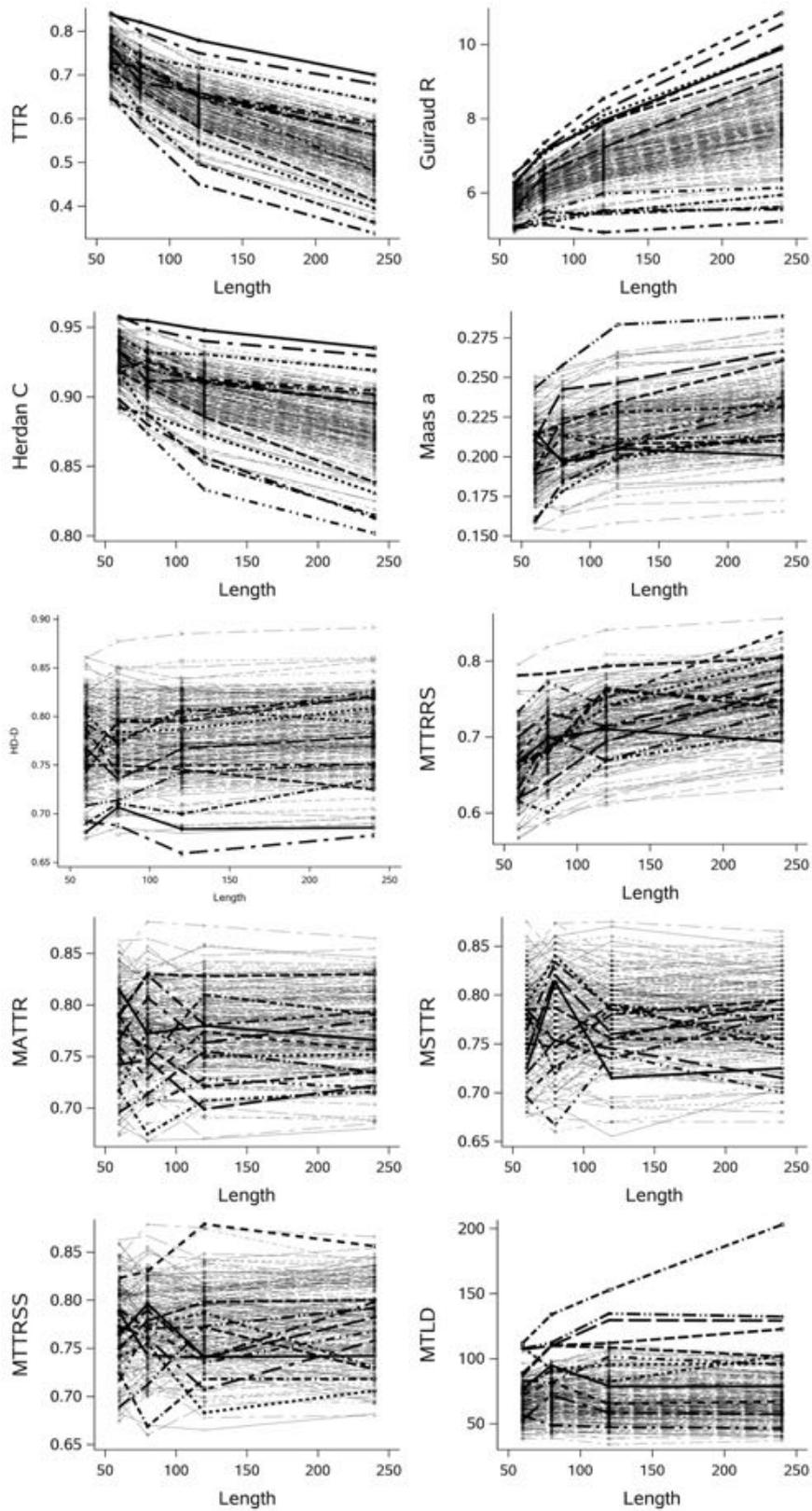

**Figure S1.2**

*Profile Graphs for ICNALE in the Random Sampling Method*

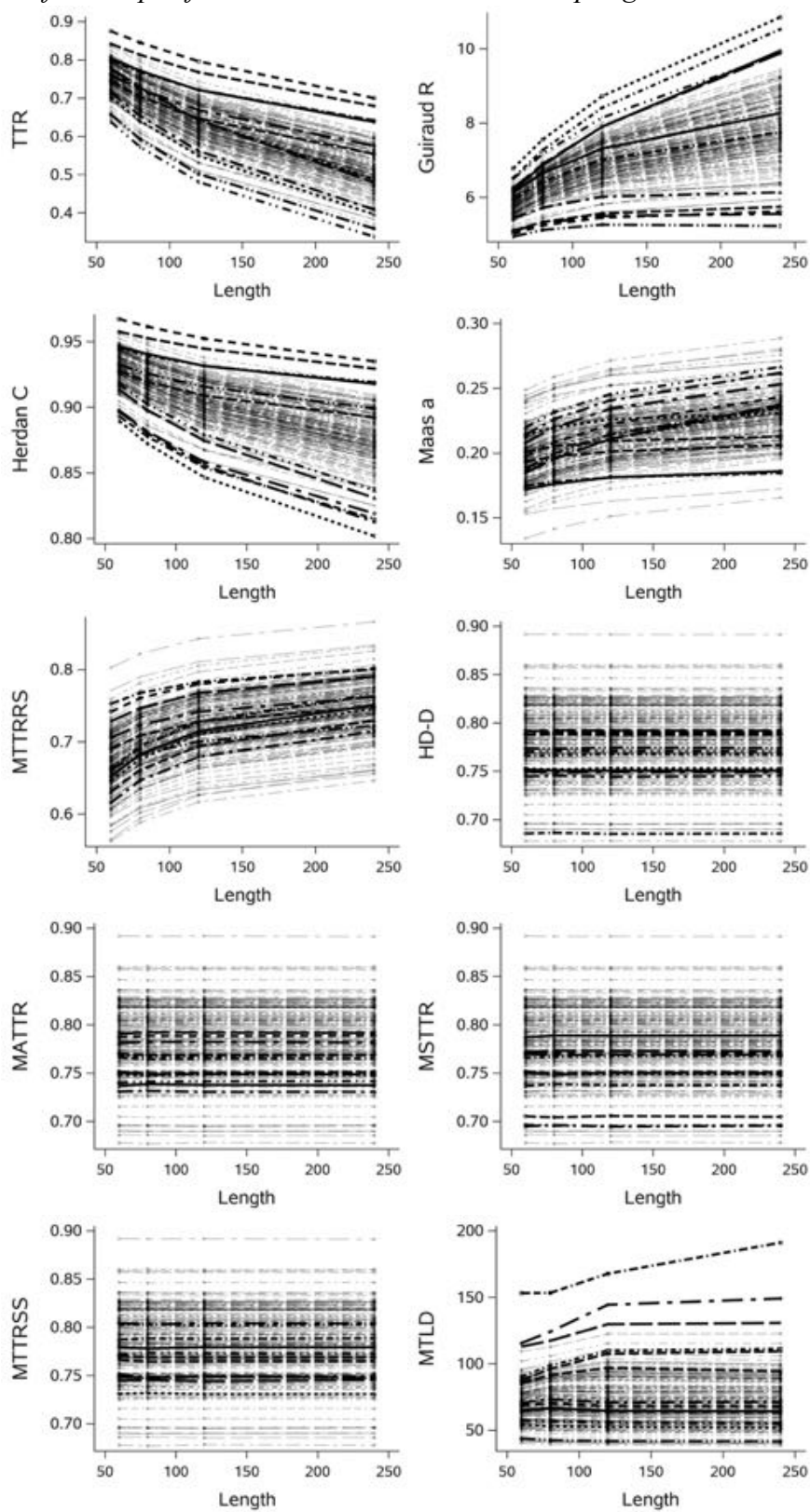

**Figure S1.3**
*Profile Graphs for ICNALE in the Ordered Random Sampling Method*

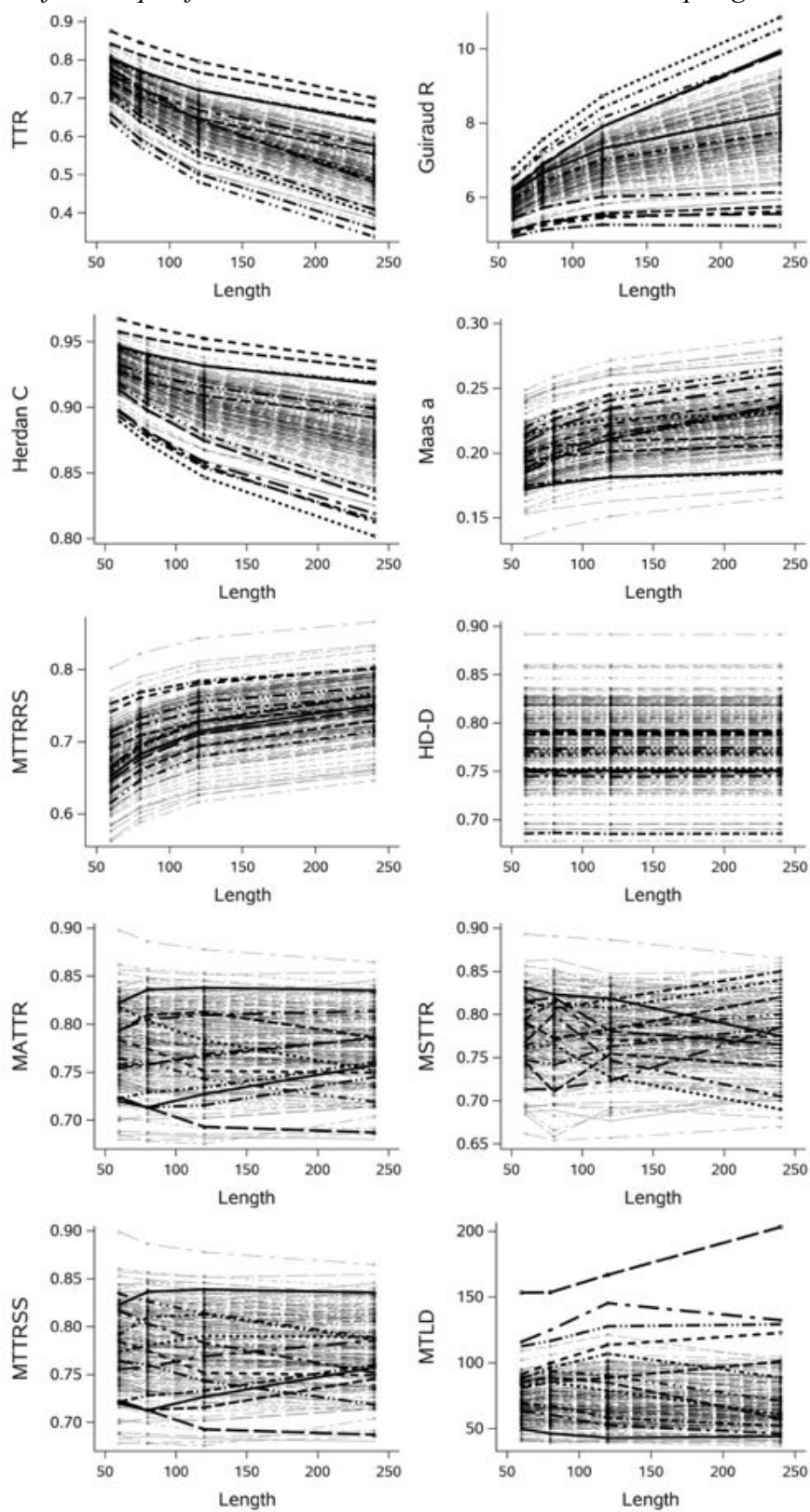

**Figure S1.4**

*Profile Graphs for ICLE in the Parallel Sampling Method*

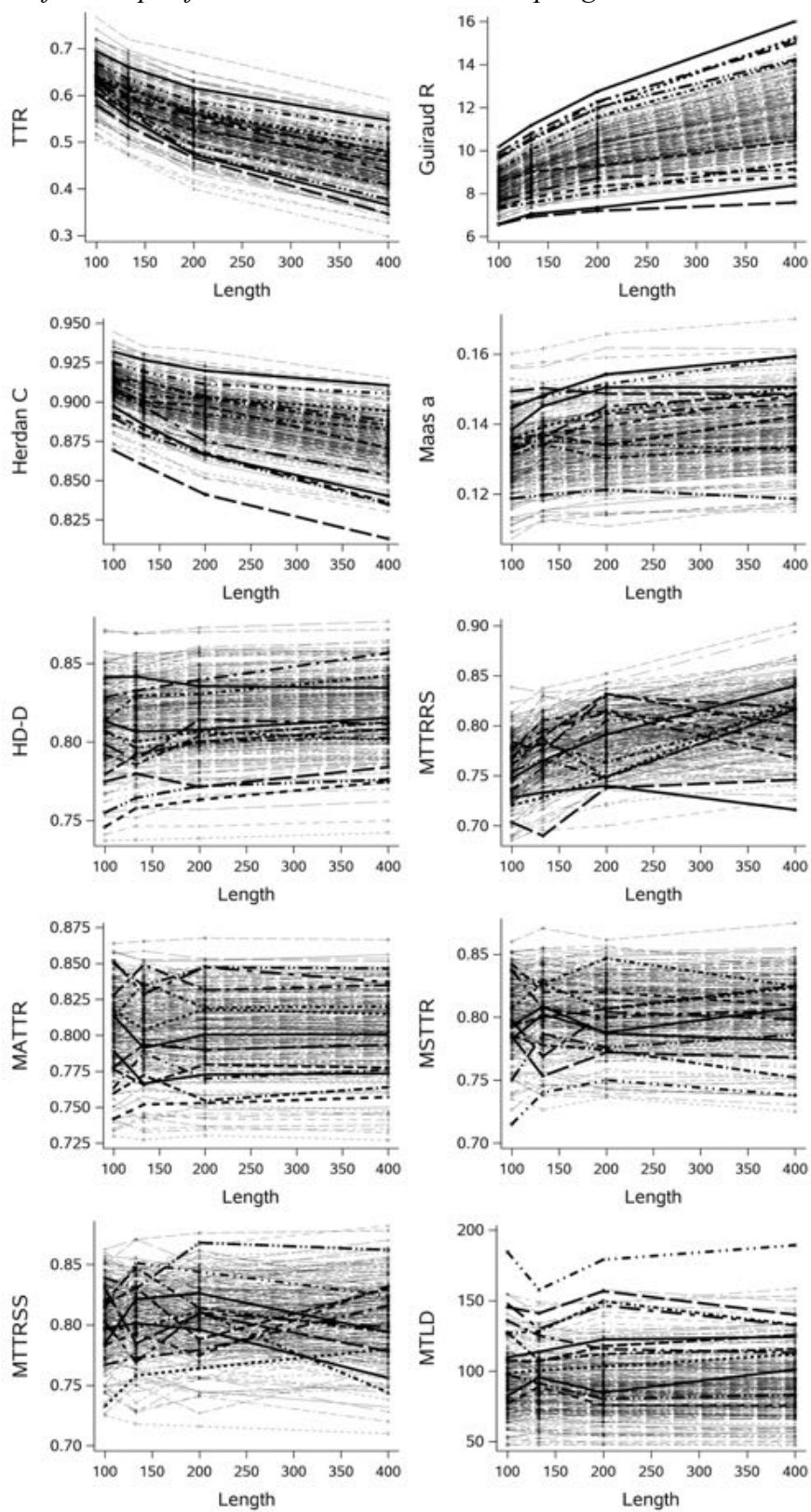

**Figure S1.5**
*Profile Graphs for ICLE in the Random Sampling Method*

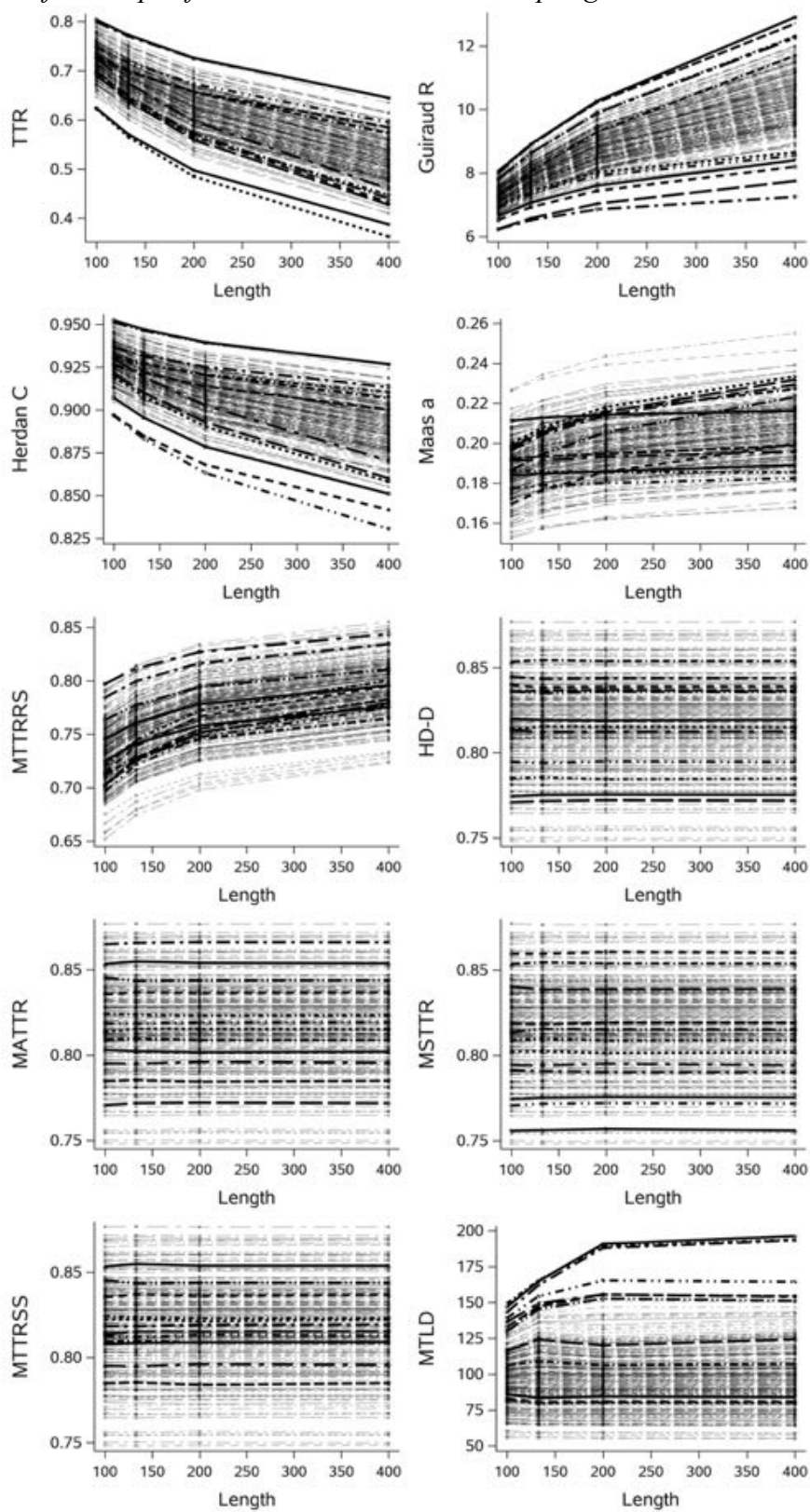

**Figure S1.6**

*Profile Graphs for ICLE in the Ordered Random Sampling Method*

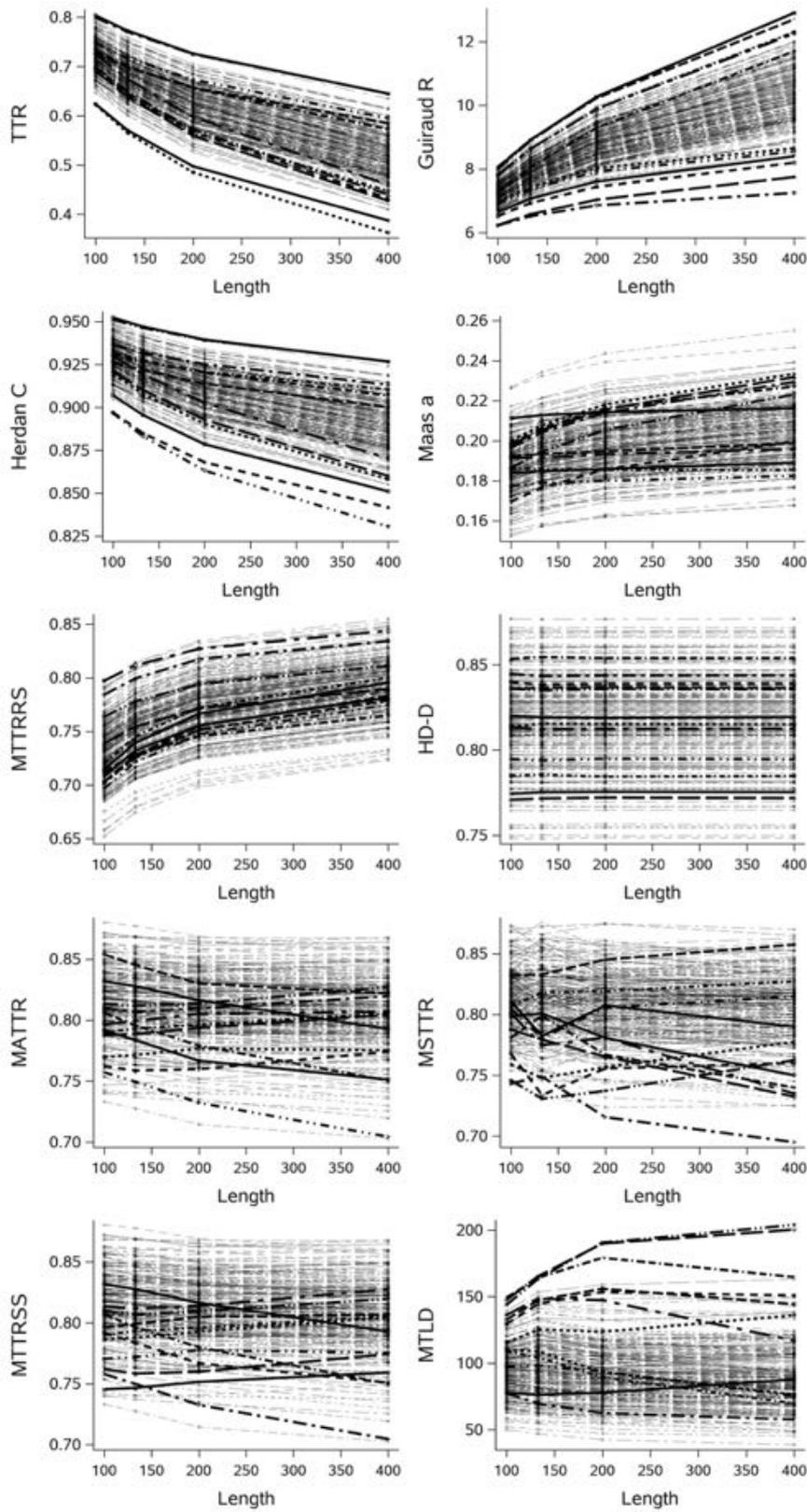

**Figure S1.7**

*Profile Graphs for ICLE in the Alternating Token Sampling Method*

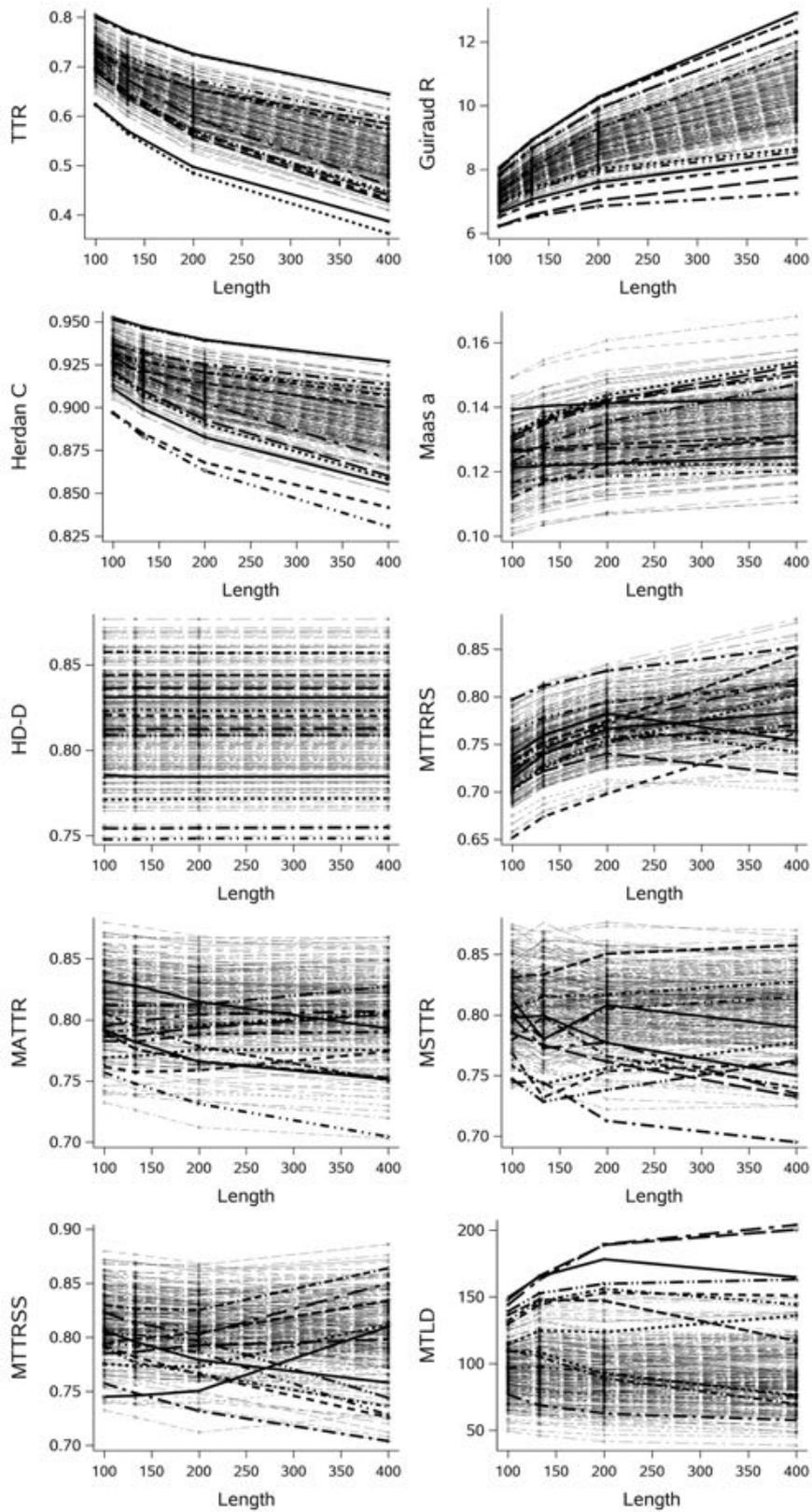

## Appendix S2: Weight of Each Token in an LD Index

The weight represents the impact that a given token has on an LD score. For example, in the TTR, each token is counted once and only once in the calculation. All tokens have thus the same weight, which is equal to *1/N*, in which *N* is the number of tokens.

For MSTTR, we have the same result only if *N* is a multiple of *n*, the length of a segment. If it is not the case, a certain number of tokens are not taken into account. If we represent by *mod(x,y)* the modulo operation, which returns the remainder of a division of *x* by *y*, the number of tokens not taken into account, is equal to *Mod(N,n)*n*. These remaining tokens received a weight of 0 while the others received a weight of *1/(N-Mod(N,n)*n)*. When *N* and *n* are unknown, we can only say that the remainder can go from *0* to *n-1* and that all these cases are equiprobable. The expectation of this number is *(n-1)/2*. Thus, for a segment length of *n* tokens, we can estimate that on average *(n-1)/2* tokens will have a weight of zero. Whether these tokens are frequent, rare or the only occurrence of a type in the whole text, they do not affect any calculated TTR. There is therefore an impact of the weight of a token on the lexical diversity score.

For MTTRSS, all tokens between positions *1* and *N-n+1* have the same probability of starting a segment while tokens occupying positions *N-n+2* to *N* cannot start a segment because it would be incomplete. A token at position *i* only intervenes in a given segment if the initial token of this segment is at one of the positions from *i-n+1* to *i* inclusive. It follows that:

- The first token will only be used when this first token is selected as the starting point. This will occur with a probability of *1/(N-n+1)*. The last token will be used only when the token in position *N-n+1* is selected as starting point.
- The second token will be used only when the first token or the second one is selected as starting point. This will happen with a probability of *2/(N-n+1)*. This probability also applies to the penultimate token.
- The same reasoning applies to the tokens that occupy positions *3* to *n-1* and *N-n+2* to *N-2*. For example, the probability that the token in position *n-1* occurs in a segment is *(n-1)/(N-n+1)*. It is also the probability that the token in position *N-n+2* occurs in a segment.
- All the tokens occupying the positions going from *n* to *N-n+1* have a probability to occur in a segment of *n/(N-n+1)*.

The weight of the tokens in the MTTRSS calculation is proportional to this probability.

In the case of MATTR, it is not correct to talk about the probability of intervening in a segment because the procedure is deterministic. On the other hand, it possible to quantify the frequency with which a token will be used according to its position in the sequence, the weight being proportional to this frequency. The first token occurs only in one sequence, the one that begins with it. The second token occurs only in two sequences, those which begin with it and with the first. This reasoning applies to the tokens occupying the positions *3* to *n-1*. From the position *n* until the position *N-n+1*, the tokens occur in *n* sequences. From position *N-n+2* to *N*, the number of sequences in which a token occurs decreases linearly exactly as it increases linearly for the *n* first tokens of the text. For example, the last token occurs in only one sequence, the one starting at position *N-n+1*. This reasoning is illustrated in Table S2.1 which gives for *N=10* and *n=4* the number of segments in which a token in position *i* can occur in the calculation of MATTR.

The impact of the weight of a token on the MATTR score, which is proportional to the number of windows in which this token is present, can be illustrated by the following example that uses the approach proposed by Covington and McFall (2010) to discuss the properties of MATTR. Let us consider a sequence of ten tokens composed of a hapax and a single other word that is present nine times. The length of the window is set to four.

Depending on the position of the hapax, the MATTR score varies from .286 to .393, as shown in Table S2.2. The only possible explanation for the differences in the scores is that the hapax does not always receive the same weight in the MATTR score.

**Table S2.1**

*Number of segments in which a given token occurs in the calculation of MATTR for N=10 and n=4*

| Position | 1 | 2 | 3 | 4 | 5 | 6 | 7 | 8 | 9 | 10 |
|---|---|---|---|---|---|---|---|---|---|---|
| Presence in a segment | x | x | x | x | | | | | | |
| | | x | x | x | x | | | | | |
| | | | x | x | x | x | | | | |
| | | | | x | x | x | x | | | |
| | | | | | x | x | x | x | | |
| | | | | | | x | x | x | x | |
| | | | | | | | x | x | x | x |
| Number of segments in which the token occurs | 1 | 2 | 3 | 4 | 4 | 4 | 4 | 3 | 2 | 1 |

Table S2.2: Computation of the MATTR scores for the ten sequences

| Line | Sequence | MATTR score |
|---|---|---|
| 1 | ABBBBBBBBB | ((2/4)*1+(1/4)*6)/7 = .286 |
| 2 | BABBBBBBBB | ((2/4)*2+(1/4)*5)/7 = .321 |
| 3 | BBABBBBBBB | ((2/4)*3+(1/4)*4)/7 = .357 |
| 4 | BBBABBBBBB | ((2/4)*4+(1/4)*3)/7 = .393 |
| 5 | BBBBABBBBB | ((2/4)*4+(1/4)*3)/7 = .393 |
| 6 | BBBBBABBBB | ((2/4)*4+(1/4)*3)/7 = .393 |
| 7 | BBBBBBABBB | ((2/4)*4+(1/4)*3)/7 = .393 |
| 8 | BBBBBBBABB | ((2/4)*3+(1/4)*4)/7 = .357 |
| 9 | BBBBBBBBAB | ((2/4)*2+(1/4)*5)/7 = .321 |
| 10 | BBBBBBBBBA | ((2/4)*1+(1/4)*6)/7 = .286 |

It is important to note that this example does not simply show that MATTR is sequence sensitive, which it is, because moving the hapax in the middle part of the sequence (lines 4 to 7) has no impact. It is only when the hapax is at the beginning or at the end of the sequence that an impact is observed; that is, when it receives a different weight. It is technically possible to build a MATTR index that allocates the same weight to all the tokens. To accomplish this, it is sufficient to use the wrap-around procedure of MTLD-W (Vidal & Jarvis, 2021) and thus to complete the incomplete windows at the end of the text using the tokens that are at the beginning of it. However, whether joining the end of a text to the beginning is more respectful of the sequential nature of a text than the global approach is questionable.

Since, to my knowledge, the problem encountered by MATTR is not highlighted in the existing literature, it is useful to formulate it also as follows. As the name suggests, MATTR is based on a moving average procedure. It is well known that this procedure does not allow to obtain a score for all points of a sequence. There are several formulations of this limitation depending on which point in the moving window the average value is assigned to: the last one as proposed by Covington and McFall (2010) or the middle one as it is frequently the case in time series. If the score is assigned to the last token, Covington and McFall (2010,

p. 96) observes that "MATTR yields a value for every point in the text except for those less than one window length from the beginning." When the average of the obtained scores is computed, the first tokens of the text are not taken into account in the same way as the following ones. These tokens have therefore a different weight in the final score. If the score is assigned to the middle token, MATTR lacks information at the beginning and at the end of the sequence: "Moving averages do not allow estimates of $f(t)$ near the ends of the time series (in the first $k$ and last $k$ periods)" (Hyndman, 2011, p. 867). The closer to the beginning (or to the end) of the sequence, the more information is missing. It is this lack of information that is quantified by the token weight.

Appendix S3: Additional Information for Table 2

**Table S3.1**

*Means, Standard Deviations and Confidence Limits for the Four Lengths for the Parallel and Random Sampling Methods*

| | Parallel Sampling | | | Random Sampling | | |
|---|---|---|---|---|---|---|
| Sample length | *M* | *SD* | CI | *M* | *SD* | CI |
| 60 | .7740 | .037 | [.7687, .7794] | .7803 | .035 | [.7752, .7854] |
| 80 | .7744 | .037 | [.7691, .7798] | .7803 | .035 | [.7752, .7854] |
| 120 | .7750 | .036 | [.7698, .7800] | .7802 | .035 | [.7751, .7853] |
| 240 | .7753 | .036 | [.7701, .7804] | .7802 | .035 | [.7751, .7853] |

The 95% confidence intervals for the ANOVA effect size, partial eta$^2$, are as follows:

Parallel Sampling: partial *eta$^2$* = .007, [.000, .031]

Random Sampling: partial *eta$^2$* = .058, [.003, .121]

Appendix S4: Procedure for Selecting 12 Profiles in the Graphs

1. Calculate the differences in the scores between the conditions taken two by two for each text.
2. For each text, count the number of times a difference is among the four largest differences or among the four smallest differences.
3. Select
- the four texts that have most of the largest differences,
- of those that remain, select the four texts that have most of the small differences, and
- of those that remain, the select four texts that have the most extreme differences (large and small).

Appendix S5: Results of Steiger's (1980) Test

Table S5.1 shows the full results of Steiger's (1980) test for two non-independent correlations that was used to determine whether there was a statistically significant difference between the largest and smallest correlation of each index with the text quality. The limits of the 95% confidence interval for the difference were obtained using Zou's (2007) procedure.

**Table S5.1**

*Comparison of the Largest and Smallest Correlations between an Index and the Text Quality for ICNALE and ICLE*

| Index | Largest $r$ | Smallest $r$ | $t(185)$ | $p$ | Lower CI | Upper CI |
|---|---|---|---|---|---|---|
| | | | ICNALE | | | |
| HDD | .342 | .285 | 1.45 | .1474 | -.019 | .132 |
| MATTR | .391 | .341 | 0.87 | .3863 | -.061 | .159 |
| MSTTR | .385 | .321 | 1.01 | .3119 | -.059 | .188 |
| MTTRSS | .373 | .299 | 1.18 | .2383 | -.048 | .196 |
| MTLD | .320 | .242 | 3.18 | .0017 | .029 | .127 |
| | | ICLE N = 223 Df = 120 | | | | |
| Index | Largest $r$ | Smallest $r$ | $t(220)$ | $p$ | Lower CI | Upper CI |
| HDD | .515 | .480 | 1.43 | .1537 | -.013 | .082 |
| MATTR | .474 | .445 | 1.55 | .1234 | -.008 | .066 |
| MSTTR | .501 | .417 | 1.97 | .0506 | -.000 | .168 |
| MTTRSS | .471 | .294 | 3.35 | .0009 | .072 | .283 |
| MTLD | .461 | .436 | 1.25 | .2113 | -.014 | .064 |

Appendix S6: Analysis of English Learners' Monologues

The main objective of the study was to determine the extent to which LD indices are sensitive to the length of the texts to which they are applied. Empirical evaluations were performed on two datasets that were relatively different in terms of text length and the participants' L1s. Both datasets consistently led to the same conclusions. This result was expected because the study focused on the logical-mathematical properties of indices and evaluation procedures. Nevertheless, the selected datasets both contained written essays. To further support the generality of the findings, this appendix presents the full analysis of a third dataset that consisted of oral productions. The conclusions reached via the analyses were identical to those reported in the main text.

**Dataset**

The spoken data used for these analyses were taken from the Corpus of English as a Foreign Language (COREFL: corefl.learnercorpora.com, version 1.0 - September 2021), which is freely available under a Creative Commons license (CC BY-NC-ND 3.0 ES). Compiled since 2012 at the University of Granada, it includes written and spoken texts from Spanish and German learners of English and from native English speakers (Lozano, Díaz-Negrillo, & Callies, 2021). The analyzed sub-corpus consists of monologues that were recorded while the English learners were asked to retell a short clip from Charles Chaplin's film *The Kid* (available at https://www.youtube.com/watch?v=eO1HvF2G2Sw). The audio files were transcribed orthographically and were tokenized by the team that collected them. The pre-processing consisted of removing the filled pauses, such as "uh" or "um", and false starts, such as " in= inside". All the texts containing at least 240 tokens were analyzed. The average length was 417 tokens (SD = 104, min = 261, max = 729). All the learners included in this dataset answered the Oxford Quick Placement Test to evaluate their levels of English proficiency.

**Analyses and results**

The analyses were identical to those described in the main text.

**First Problem: Evaluation of the Text Length Sensitivity**
*ICC Analysis and Graphical Representation of the Profiles*
Figure S6.1 shows the ICCs for the four evaluation methods and the ten indices. The mean ICC is represented by a circle, while the bars represent the 95% confidence interval for the ICC population values. As explained in the main text, the parallel sampling method is problematic and its conclusions are not reliable. Figures S6.2 to S6.5 show the impact of length on the indices using the LD text profiles.

TTR, Guiraud, Herdan, Maas, and MTTRSS clearly showed a length bias. The profiles for MATTR, MSTTR, MTTRSS, and MTLD did not show any bias, but there were important crossings, which explained why the ICC was sometimes quite far from 1. Only the profiles for HD-D were almost perfectly stable.

*Second Problem: Impact of the Index Parameter*
Figure S6.6 shows the consistency of the indices as a function of the parameter value measured by the ICC. MTLD and HDD were less affected, followed by MATTR. Figure S6.7 illustrates the impact of parameter manipulation on the LD scores. In this figure, the scores for each index for each value of the parameter were centered on 0 in order to eliminate bias and to match the data used for the ICC precisely when it measures consistency. For all the indices, the LD score of some texts increased with the increase in the parameter value, while this was the opposite for other texts and produced profile crossings.

**Figure S6.1**
*ICCs for Text Length Sensitivity for COREFL*

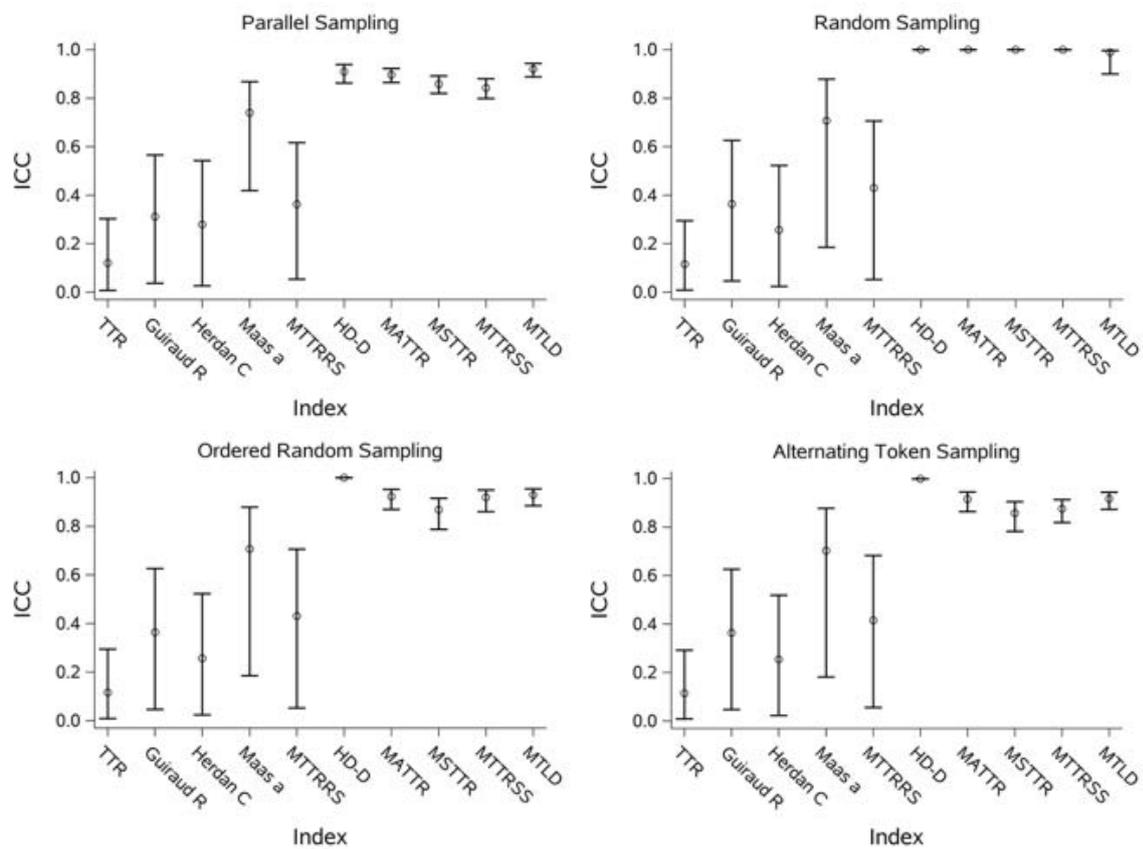

**Figure S6.2**

*Profile Graphs for COREFL in the Parallel Sampling Method*

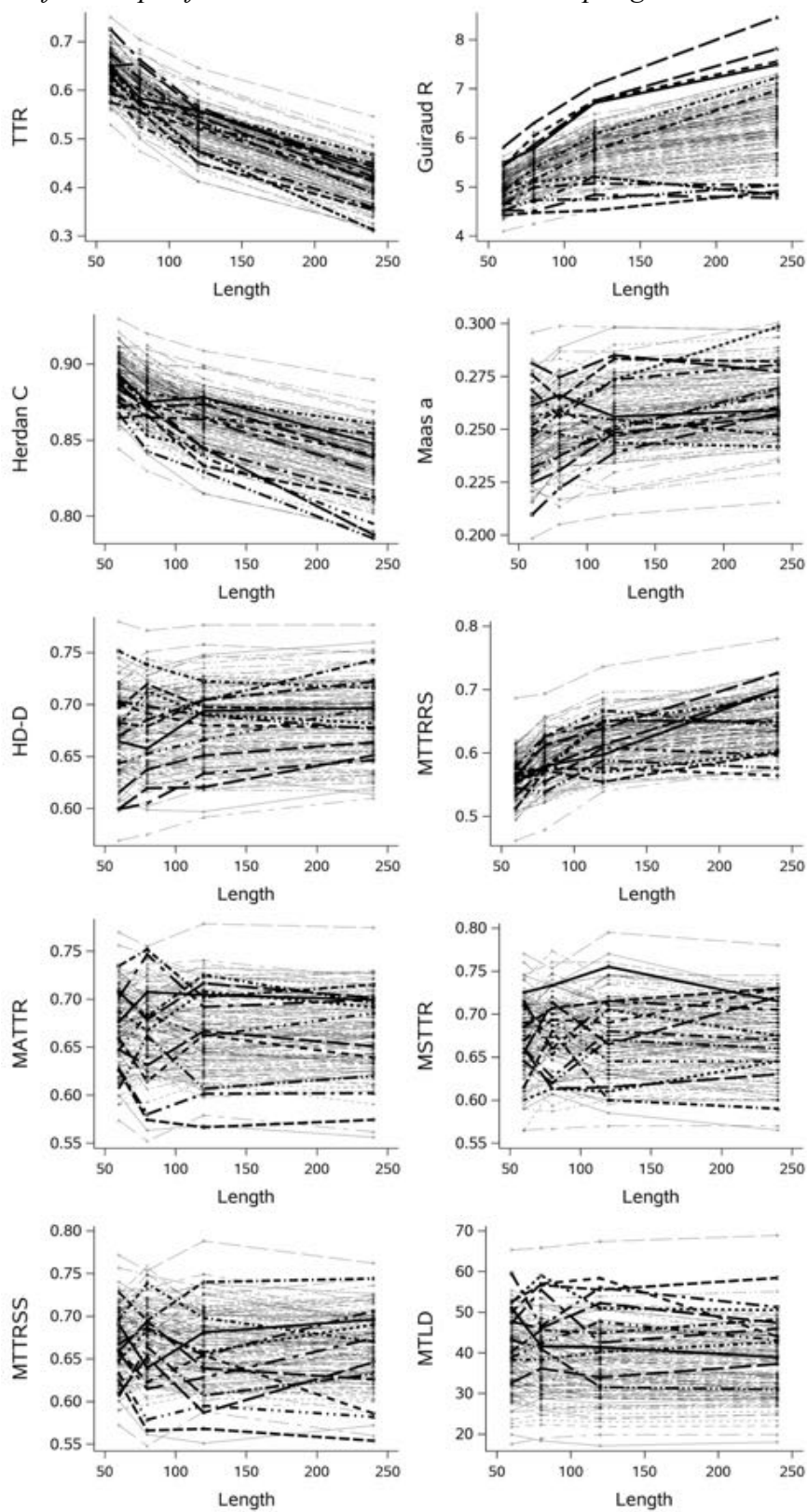

**Figure S6.3**
*Profile Graphs for COREFL in the Random Sampling Method*

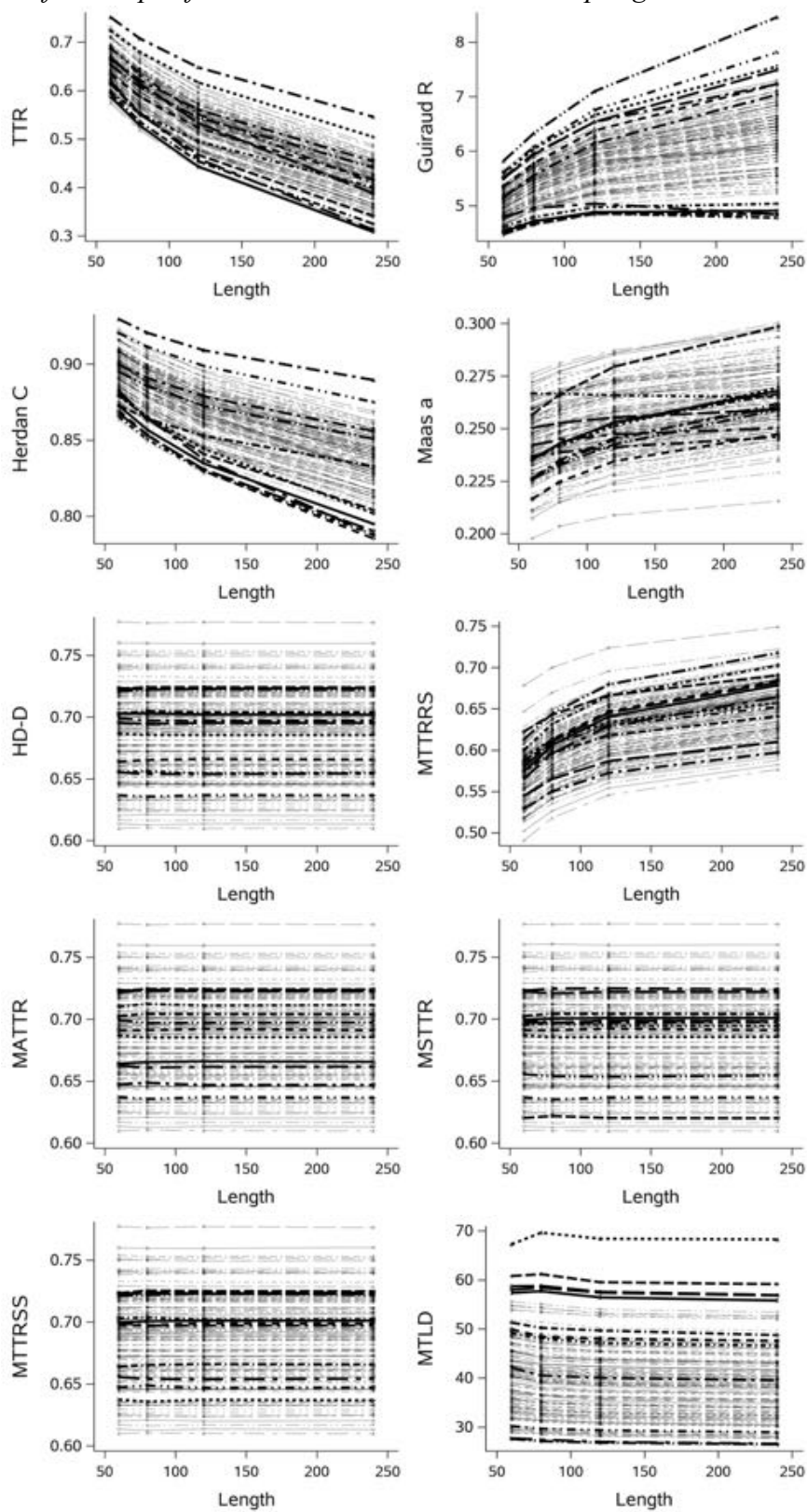

**Figure S6.4**
*Profile Graphs for COREFL in the Ordered Random Sampling Method*

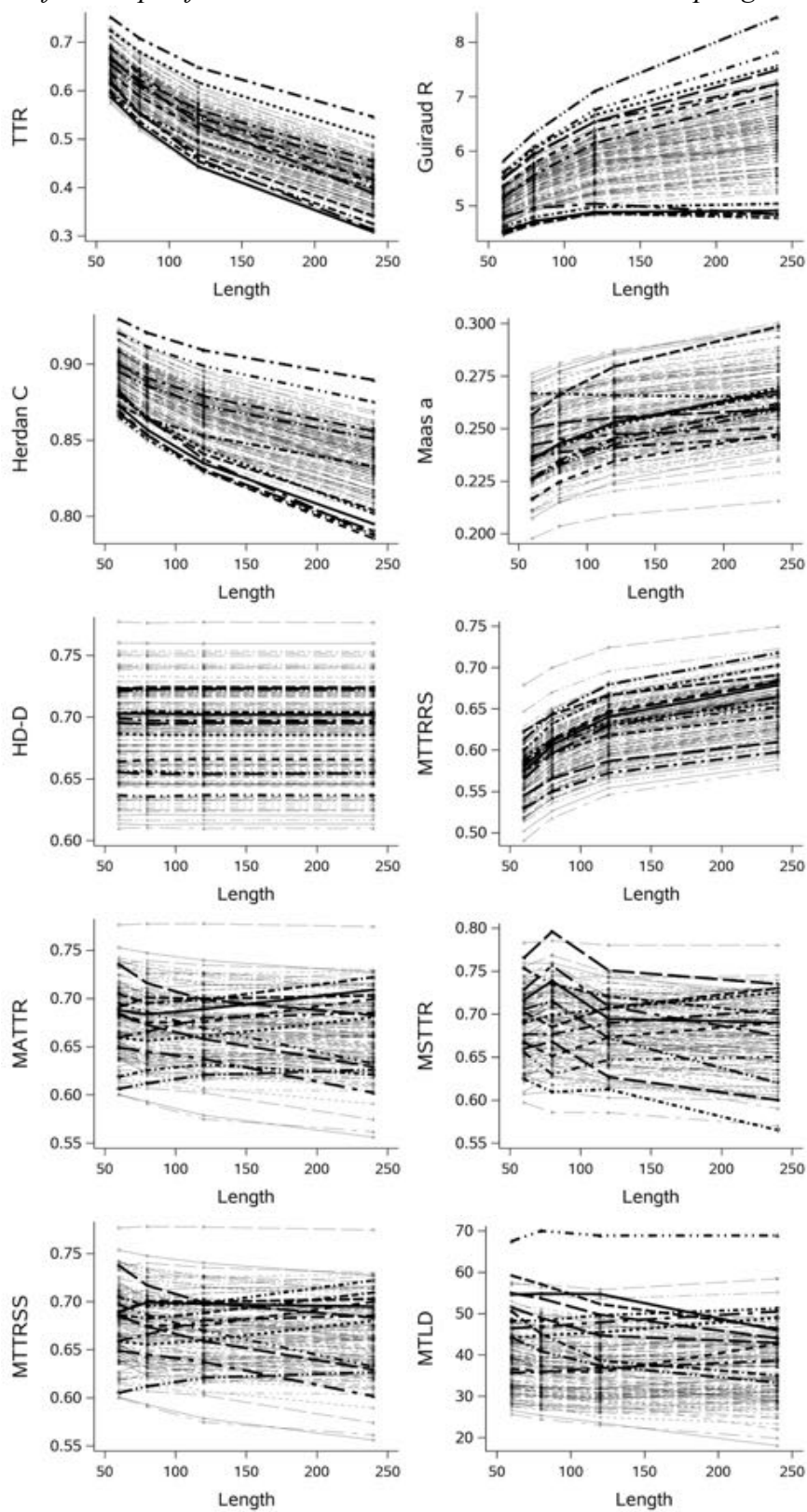

**Figure S6.5**

*Profile Graphs for COREFL in the Alternating Token Sampling Method*

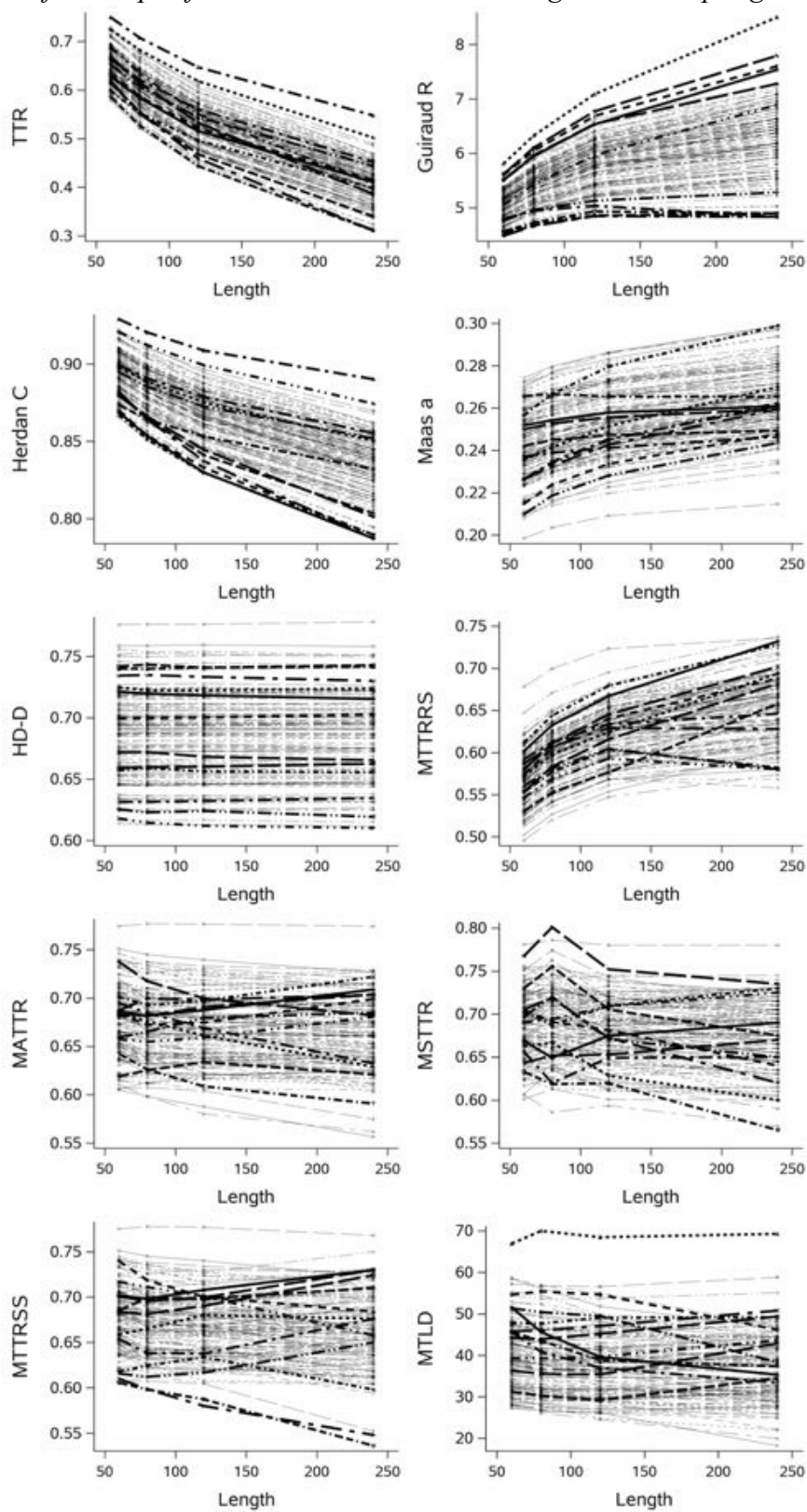

**Figure S6.6**
*ICCs for the Impact of the Index Parameter for COREFL*

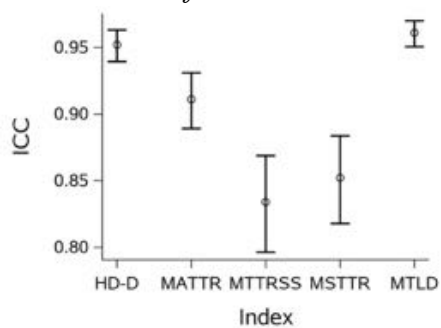

**Figure S6.7**
*Profile Graphs for the Parameter Impact for COREFL*

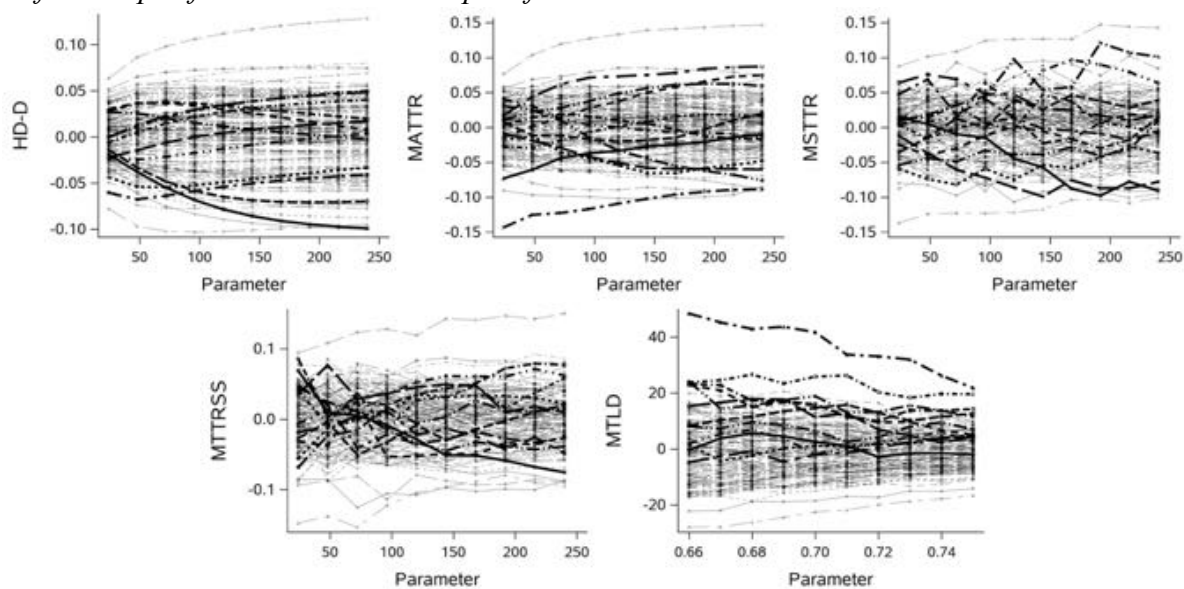

*Impact on the Correlation between LD and English Proficiency level*
Table S6.1 shows the correlation between the LD score and the learner's level of English proficiency for each parameter value. The largest value for each index is in bold, and the smallest is underlined. As shown in Table S6.2, none of the differences between the largest and smallest correlations were statistically significant according to Steiger's (1980) test.

**Table S6.1**
*Correlations between the LD and the Text Quality for COREFL for Different Parameter Values*

| Length | HD-D | MATTR | MSTTR | MTTRSS | Factor | MTLD |
|--------|------|-------|-------|--------|--------|------|
| 24 | .358 | .374 | .313 | .306 | .66 | .300 |
| 48 | .394 | .375 | .279 | .316 | .67 | .306 |
| 72 | .408 | .370 | .343 | .370 | .68 | .330 |
| 96 | .412 | .382 | .311 | .374 | .69 | .299 |
| 120 | **.413** | **.385** | .312 | .380 | .70 | .311 |
| 144 | .411 | .383 | .304 | **.402** | .71 | .313 |
| 168 | .410 | .377 | **.346** | .365 | .72 | .331 |
| 192 | .408 | .371 | .314 | .352 | .73 | .321 |
| 216 | .406 | .358 | .330 | .354 | .74 | .314 |
| 240 | .404 | .335 | .336 | .309 | .75 | **.332** |

**Table S6.2**

*Comparison of the Largest and Smallest Correlations between an Index and the Text Quality for COREFL*

| | CORELF | | | | | |
|---|---|---|---|---|---|---|
| Index | Largest *r* | Smallest *r* | *t*(127) | *p* | Lower CI | Upper CI |
| HDD | .413 | .358 | 1.58 | .117 | -.013 | .123 |
| MATTR | .385 | .335 | 1.85 | .066 | -.003 | .103 |
| MSTTR | .346 | .279 | 1.25 | .213 | -.038 | .172 |
| MTTRSS | .402 | .306 | 1.43 | .154 | -.035 | .228 |
| MTLD | .331 | .299 | 1.16 | .249 | -.022 | .086 |

Conclusion

The analysis of this third dataset confirmed the conclusions for the two written datasets presented in the main text.

Appendix S7: Obtaining the Main LD Indices by Means of Two Freely Available Tools

This appendix explains how to obtain the main LD indices that were analyzed in this paper by means of two freely available tools, TAALED and the koRpus package. It also shows that the indices calculated by these tools were identical to those calculated by the SAS program used in this paper. This SAS program is included at the end.

TAALED (Kyle et al., 2021) is standalone software that is designed to calculate LD indices. It is controlled through an easy-to-use window interface.

The koRpus package (Michalke, 2020) works with R software. A text can be analyzed using the following commands.

```
-----
install.packages("koRpus")
install.koRpus.lang(c("en"))
library("koRpus")
library(koRpus.lang.en)
setwd("/Users/c/Desktop/Example")
tagged.text.obj <-
tokenize("./AliceNoun2.txt",lang="en",detect=c(parag=TRUE, hline=TRUE))
TTR<-TTR(tagged.text.obj)
TTR@TTR
R<-R.ld(tagged.text.obj)
R@R.ld
C<-C.ld(tagged.text.obj)
C@C.ld
maas<-maas(tagged.text.obj)
maas@Maas
HDD<-HDD(tagged.text.obj)
HDD@HDD
MATTR<-MATTR(tagged.text.obj,window=50)
MATTR@MATTR
MSTTR<-MSTTR(tagged.text.obj,segment=50)
MSTTR@MSTTR
MTLD<-MTLD(tagged.text.obj,MA=FALSE)
MTLD@MTLD
-----
```

As an example, I used the beginning of Lewis Caroll's *Alice in Wonderland*, which is the text that was used as the main example by Baayen (2001), and stopped the excerpt after the first 100 types, or 165 tokens. The tokenized extract is presented below.

*alice was beginning to get very tired of sitting by her sister on the bank and of having nothing to do once or twice she had peeped into the book her sister was reading but it had no pictures or conversations in it and what is the use of a book thought alice without pictures or conversations so she was considering in her own mind as well as she could for the hot day made her feel very sleepy and stupid whether the pleasure of making a daisy-chain would be worth the trouble of getting up and picking the daisies when suddenly a white rabbit with pink eyes ran close by her there was nothing so very remarkable in that nor did alice think it so very much out of the way to hear the rabbit say to itself oh dear oh dear i shall be late when she thought it over afterwards it occurred to her that she ought to have wondered at this*

The main difficulty when comparing several types of text analysis software is that they frequently perform a re-tokenization of the text, which includes the modification of some tokens and even the deletion of others. As was the case for this extract, each type was replaced by a noun that was treated in the same way by the evaluated tools. For example, *Alice* was recoded as *art* and *the* as *two*. The recoded extract is presented below.

> *art team case state friend teacher side lot president city hand power member two business back lot group life state fact minute moment student point girl name home two change hand power team party child house girl law number moment country history house back war hour two system lot air change service art word number moment country program point team company history hand mother kid body time body point day force two head end issue hand father teacher problem back question way two others lot job air education year car world two story lot game study back night two door water reason air week parent woman office family part community city hand right team life program teacher people history research level face art room house program teacher kind month lot two thing state guy two parent person state idea man eye man eye health place car information water point service house morning area house line state hand research point money state government work book school*

Table S7.1 provides the main indices that were computed using this extract by the three approaches. These indices are identical except for that of Maas. The reason is that TAALED calculates $a^2$ while koRpus and the SAS program calculate $a$. The base 10 is used for the logarithm, as in Maas (1972).

**Table S7.1**
*LD Indices Produced by Three Types of Software*

|         | Token | Type | TTR   | Guiraud | Herdan | Maas  | HD-D  | MATTR | MSTTR | MTLD    |
|---------|-------|------|-------|---------|--------|-------|-------|-------|-------|---------|
| TAALED  | 165   | 100  | .6061 | 7.7850  | .9019  | .0442 | .8278 | .7993 | .8133 | 77.8337 |
| koRpus  | 165   | 100  | .6061 | 7.7850  | .9019  | .1386 | .8278 | .7993 | .8133 | 77.8337 |
| SAS     | 165   | 100  | .6061 | 7.7850  | .9019  | .1386 | .8278 | .7993 | .8133 | 77.8337 |

Note: HD-D is called ATTR in koRpus.

The SAS program for this example is available through the following OSF repository: https://osf.io/5xpcw/ (Cr01LexDivAnaExample.sas).

Appendix S8: SAS Code and R Function Used

The SAS code and the R function used in this study are available through the following OSF repository: https://osf.io/5xpcw/ ( SASCodeAndR.zip). The following programs are included.

For the example:
 Cr01LexDivAnaExample.sas

For the experiments:
The programs take an SAS dataset containing one observation per token with a variable that identifies the text and another that indicates the order of the tokens in the text as input.
As an example, Cr11ICNALERead.sas reads all the tokenized ICNALE files in a folder and puts them in this format

*Calculate the indices for the four methods
 Cr21LexDivGenFileAllMethods.sas
  Output for Parallel Sampling: zIceMeth_pasa
  Output for Random Sampling and Ordered Random Sampling: zIceMeth_rnd2
  Output for Alternating Token Sampling: zIceMeth_splitr

*Calculate the indices for the parameter analysis
 Cr22LexDivGenFileParameters.sas
  Output for HDD: zIceParam_hdd
  Output for MATTR: zIceParam_mattr
  Output for MSTTR: zIceParam_msttr
  Output for MTTRSS: zIceParam_mttrss
  Output for MTLD: zIceParam_mtld

*Output files to calculate ICCs in R for the methods
 Cr31LexDivMethodsOutForICC.sas
  Output: a txt file for each method and each index named zicemeth_pasa_ttr.txt (for example)

*Output files to calculate ICCs in R for the parameters
 Cr32LexDivParametersOutForICC.sas
  Output: a txt file for each index named zIceParam_hdd.txt (for example)

*Calculate ICCs in R for the methods
 MyICCMeth.R
  Output: ICCResMeth.txt

*Calculate ICCs in R for the parameters
 MyICCParam.R
  Output: ICCResParam.txt

*ICC graphs for the methods
 Cr41LexDivMethodsGraphICC.sas

*Profile graphs for the methods
 Cr42LexDivMethodsGraphIdx.sas

*ICC graphs for the parameters
 Cr43LexDivParametersGraphICC.sas

*Profile graphs and correlations for the parameters
 Cr44LexDivParametersGraphIdx.sas

*Program for Figure 6
 Cr51SimulHdd.sas

Appendix S9: Implementing the Sampling Methods in Python

In order to facilitate the replication of the analyses, but also the application of the proposed sampling methods to other datasets (L1, longer or shorter texts...), this appendix provides two Python scripts, available through the following OSF repository: https://osf.io/5xpcw/ (PythonScripts.zip), that implement the random sampling method, the ordered random sampling method and the alternating token sampling method.

*Rand_OrdRand.py* implements the random sampling method and the ordered random sampling method so that the same samples of tokens are analyzed by both methods. The only difference is that the tokens are replaced in their original order before the indices are computed for the ordered random sampling method but not for the random sampling method. To do so, it is not the tokens in the text that are permuted, but an index vector containing integers ranging from 0 to the number of tokens in the text minus 1 (in Python, indexing starts at 0). To obtain a sample of *m* tokens using the random sampling method, the tokens in the text at the first *m* positions of the permuted index vector are extracted. For the ordered random sampling method, the same tokens are extracted, but they are reordered according to the indexes.

*Alternating.py* implements the alternating token sampling method which generalize the split-half procedure proposed by McKee et al. (2000). This method constructs samples containing half, a third or a quarter of the tokens in the text by randomly selecting one token out of two, three or four successive tokens. To do so, the tokens are permuted independently within each successive snippet of two, three or four tokens (the *Pnsplit* parameter) and the samples are obtained by taking one token out of *Pnsplit* tokens, starting with the first, then the second, up to the *Pnsplit*-th.

These scripts read as input a .txt file containing a tokenized text (see TaaledSample.txt in the archive). They provide two types of output.

**The samples produced by each method.**
Two parameters (SHOW_INSTANCE and N_SHOW_INSTANCE) allow to obtain the samples produced by each method. To highlight the characteristics of these samples, the scripts can be applied to a pseudo-text composed of numbers from 1 to 303 (Numbers.txt in the zip file). By way of example, here are the first samples produced by the random sampling method and by the ordered random sampling method on the basis of this pseudo-text:

```
Random | Sample length = 151 | Niter = 1
['253', '72', '169', '301', '98', '95', '263', '81', '182', '275', '87', '86',
'297', '32', '281', '54', '163', '97', '108', '104', '298', '139', '93', '101',
'91', '45', '217', '242', '183', '194', '248', '254', '129', '114', '172', '110',
'197', '218', '200', '160', '211', '162', '208', '187', '165', '255', '273', '43',
'238', '245', '264', '125', '35', '29', '173', '221', '219', '156', '33', '198',
'249', '294', '116', '234', '90', '170', '246', '30', '7', '83', '178', '224',
'303', '220', '189', '196', '209', '205', '120', '150', '34', '71', '250', '47',
'4', '192', '282', '181', '144', '158', '277', '203', '113', '299', '107', '126',
'268', '223', '239', '215', '10', '293', '259', '8', '236', '240', '302', '79',
'283', '118', '143', '272', '204', '167', '166', '53', '100', '52', '140', '199',
'74', '12', '122', '149', '300', '175', '258', '288', '103', '124', '176', '60',
'216', '66', '155', '260', '41', '186', '89', '94', '230', '243', '127', '214',
'292', '287', '290', '57', '152', '190', '70']
Ordered Random | Sample length = 151 | Niter = 1
['4', '7', '8', '10', '12', '29', '30', '32', '33', '34', '35', '41', '43', '45',
'47', '52', '53', '54', '57', '60', '66', '70', '71', '72', '74', '79', '81', '83',
'86', '87', '89', '90', '91', '93', '94', '95', '97', '98', '100', '101', '103',
```

```
'104', '107', '108', '110', '113', '114', '116', '118', '120', '122', '124', '125',
'126', '127', '129', '139', '140', '143', '144', '149', '150', '152', '155', '156',
'158', '160', '162', '163', '165', '166', '167', '169', '170', '172', '173', '175',
'176', '178', '181', '182', '183', '186', '187', '189', '190', '192', '194', '196',
'197', '198', '199', '200', '203', '204', '205', '208', '209', '211', '214', '215',
'216', '217', '218', '219', '220', '221', '223', '224', '230', '234', '236', '238',
'239', '240', '242', '243', '245', '246', '248', '249', '250', '253', '254', '255',
'258', '259', '260', '263', '264', '268', '272', '273', '275', '277', '281', '282',
'283', '287', '288', '290', '292', '293', '294', '297', '298', '299', '300', '301',
'302', '303']

Random | Sample length = 101 | Niter = 1
['164', '258', '264', '261', '59', '118', '36', '222', '28', '185', '275', '260',
'287', '210', '8', '148', '217', '121', '52', '292', '282', '192', '29', '207',
'214', '45', '170', '31', '20', '278', '297', '158', '268', '169', '87', '291',
'21', '74', '76', '198', '95', '128', '152', '177', '237', '104', '155', '188',
'167', '295', '72', '129', '181', '213', '99', '252', '141', '30', '271', '3',
'56', '108', '130', '256', '126', '232', '142', '303', '143', '151', '246', '221',
'115', '228', '145', '204', '14', '176', '93', '39', '111', '84', '242', '86',
'231', '116', '83', '249', '112', '201', '103', '157', '247', '60', '131', '233',
'19', '27', '286', '190', '183']
Ordered Random | Sample length = 101 | Niter = 1
['3', '8', '14', '19', '20', '21', '27', '28', '29', '30', '31', '36', '39', '45',
'52', '56', '59', '60', '72', '74', '76', '83', '84', '86', '87', '93', '95', '99',
'103', '104', '108', '111', '112', '115', '116', '118', '121', '126', '128', '129',
'130', '131', '141', '142', '143', '145', '148', '151', '152', '155', '157', '158',
'164', '167', '169', '170', '176', '177', '181', '183', '185', '188', '190', '192',
'198', '201', '204', '207', '210', '213', '214', '217', '221', '222', '228', '231',
'232', '233', '237', '242', '246', '247', '249', '252', '256', '258', '260', '261',
'264', '268', '271', '275', '278', '282', '286', '287', '291', '292', '295', '297',
'303']

Random | Sample length = 75 | Niter = 1
['243', '32', '111', '156', '204', '301', '188', '14', '138', '13', '298', '151',
'129', '237', '164', '63', '97', '96', '287', '288', '108', '110', '15', '302',
'190', '8', '192', '274', '28', '168', '162', '73', '86', '67', '43', '139', '100',
'10', '145', '277', '35', '161', '84', '238', '265', '144', '50', '77', '226',
'90', '52', '80', '112', '158', '27', '94', '276', '117', '39', '270', '53', '184',
'150', '133', '72', '91', '38', '22', '25', '42', '292', '29', '223', '101', '259']
Ordered Random | Sample length = 75 | Niter = 1
['8', '10', '13', '14', '15', '22', '25', '27', '28', '29', '32', '35', '38', '39',
'42', '43', '50', '52', '53', '63', '67', '72', '73', '77', '80', '84', '86', '90',
'91', '94', '96', '97', '100', '101', '108', '110', '111', '112', '117', '129',
'133', '138', '139', '144', '145', '150', '151', '156', '158', '161', '162', '164',
'168', '184', '188', '190', '192', '204', '223', '226', '237', '238', '243', '259',
'265', '270', '274', '276', '277', '287', '288', '292', '298', '301', '302']
```

## The HD-D and MATTR indices

To show how HD-D and MATTR can be calculated on the basis of the samples produced by each method, the scripts includes several Python functions (with slight modifications) from the underlying code for TAALED (Kyle et al., 2021), more precisely from ld_32.py, authored by Kristopher Kyle and available at https://github.com/LCR-ADS-Lab/TAALED/blob/main/dev/ld_32.py. By adjusting a parameter, the methods can be applied to an example text of TAALED (TaaledSample.txt). Here are the results obtained for each method by performing 5,000 randomizations:

```
Random    | Sample length =  140 | HDD =  0.871083 | MATTR =  0.850942
Ordered   | Sample length =  140 | HDD =  0.871083 | MATTR =  0.840838
Random    | Sample length =   93 | HDD =  0.871010 | MATTR =  0.851485
Ordered   | Sample length =   93 | HDD =  0.871010 | MATTR =  0.850711
Random    | Sample length =   70 | HDD =  0.871239 | MATTR =  0.851522
Ordered   | Sample length =   70 | HDD =  0.871239 | MATTR =  0.850743

Alternating | Segment length =  140 | HDD =  0.870906 | MATTR =  0.839625
Alternating | Segment length =   93 | HDD =  0.870912 | MATTR =  0.850383
```

```
Alternating | Segment length =   70 | HDD =  0.871491 | MATTR = 0.850603
```

Appendix S10: In-Depth Discussion of the Most Important Studies
Claiming That HD-D is Sensitive to Text Length

The present study highlights the extremely high efficiency of HD-D for controlling the length effect. This conclusion contradicts that of several previous studies. What follows is an in-depth discussion of these most important studies.

McCarthy, P. M., & Jarvis, S. (2007). vocd: A theoretical and empirical evaluation. Language Testing, 24, 459-488. https://doi.org/10.1177/0265532207080767

McCarthy and Jarvis (2007) argued that HD-D is sensitive to text length on the basis of three arguments. First, an empirical analysis (pp. 475-476) described as follows (ATTR-42 is the name used by McCarthy and Jarvis to refer to HD-D):

> To confirm that this method of measurement truly does exhibit text-length effects in our own database, we calculated ATTR-42 for the first 50 through the first 200 tokens of each of the 141 texts in our database that are longer than 200 words in length. (We used ATTR instead of SOP because the ATTR scale, which ranges from 0 to 1.00, is easier to interpret, and we used ATTR instead of D because ATTR is more precise and consistent.) Given that it is not feasible to illustrate the ATTR-42 growth curves for all 141 texts, we have chosen instead to show a single curve representing the central tendency of these texts. This is shown in Figure 5. This figure shows that after a slight dip in ATTR-42 from about the 50th to the 100th token, the central tendency is for the ATTR-42 of texts to move gradually and steadily upward.

**Figure S10.1:** Evolution of the Mean HD-D Computed on the 188 Texts of ICNALE.

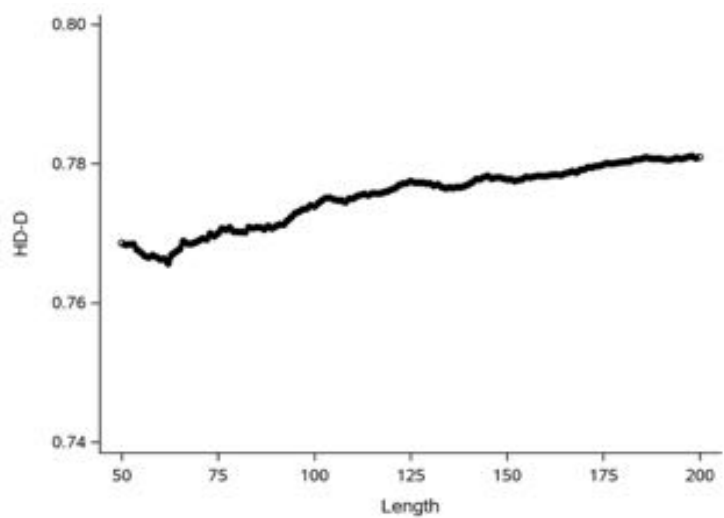

This demonstration is problematic because it relies on samples that obviously do not have the same lexical content, since the next token is added to the sample each time. If there is an evolution of lexical diversity in these texts, a length-insensitive index will detect it. To support this explanation, I performed McCarthy and Jarvis' analysis on the ICNALE dataset used in the manuscript. Figure S10.1 shows the evolution of the mean HD-D computed on the 188 texts. As in McCarthy and Jarvis' work, a small decrease is observed at the beginning of the curve, followed by an increase. Figure S10.2 shows the 12 profiles selected by the systematic procedure used in the manuscript; it shows that some profiles increased, but also that others decreased. This observation is incompatible with the assertion that HD-D increases with text length. However, it is perfectly compatible with an evolution of lexical diversity in texts, which is one phenomenon that motivated Covington and McFall (2010) to develop MATTR.

**Figure S10.2:** Twelve Individual HD-D Profiles from ICNALE.

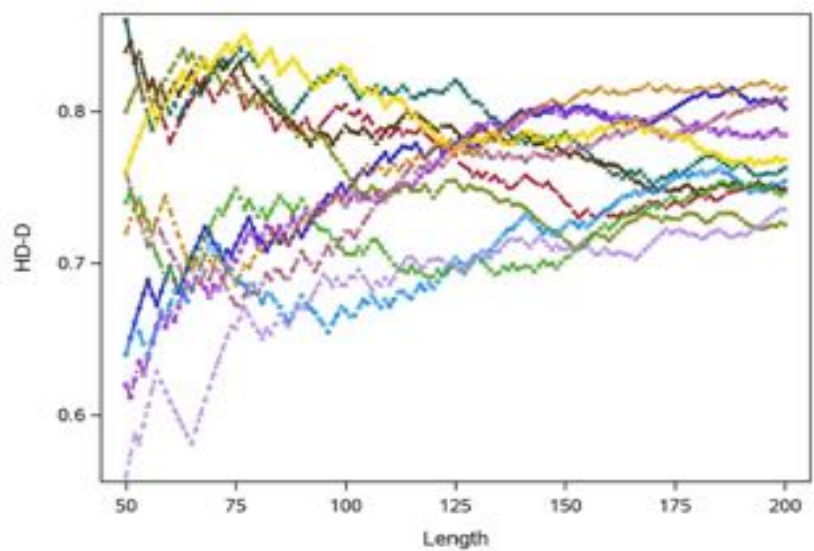

In the next part of McCarthy and Jarvis' (2007) work, which is one of the most important articles for the present study, the authors used the parallel sampling method (p. 479). As explained in the main text, these analyses cannot be used to claim that HD-D is sensitive to text length.

Finally, McCarthy and Jarvis (2007) proposed a mathematical argument by analyzing the impact of adding an occurrence of a type already present in a 100-word text on HD-D by varying the frequency of that type. The authors showed that the reduction of HD-D was increasingly lower the less frequent the type was, up to the point at which the addition of a second occurrence of a type that was only present once in the text produced an increase of HD-D and thus of lexical diversity. This observation, which the authors rightly called "surprising and profound" (p. 475) has noting to do with any sensibility to text length as explained at the end of the discussion of Figure 6 of the main text.

Fergadiotis and colleagues (2013, 2015) also concluded that HD-D is somewhat sensitive to text length. Their work is based on a different approach from the one that is almost always used in the field. Instead of trying to determine directly whether LD indices are sensitive to text length, the authors used an indirect approach based on multidimensional analysis techniques (factor analysis and structural equation modeling). These techniques are mainly aimed at determining whether different indices are manifestations of the same latent variable; that is, LD. It is well established that the results of such analyses are affected by the variables that are selected for the analysis:

One of the characteristics of factor analysis that bothers many users is the fact that what you find depends on what you decide to analyze. If one dimension is overrepresented by tests, while another is underrepresented by tests, the former dimension will be the most dominant dimension in the analysis, which may or may not have anything to do with its potency in real life settings. Many users are familiar with this property, which might be called the 'what you get out of it depends on what you put into it' phenomenon (Dorans & Lawrence, 1999, p. 6).

For example, in one of their studies, Fergadiotis et al. (2015) analyzed two local indices, MATTR and MTLD, and one global index, vocd, as well as Maas' index, which is another global index that is well known to be length sensitive (McCarthy & Jarvis, 2013; Tweedie & Baayen, 1998). Further studies are needed to assess the impact of this choice on the results but, in general, Fergadiotis and colleagues' (2013, 2015) arguments are indirect, unlike those presented in the manuscript.